\documentclass{article}

\usepackage[preprint,nonatbib]{neurips_2020}

\usepackage[utf8]{inputenc} % allow utf-8 input
\usepackage[T1]{fontenc}    % use 8-bit T1 fonts
\usepackage{hyperref}       % hyperlinks
\usepackage{url}            % simple URL typesetting
\usepackage{booktabs}       % professional-quality tables
\usepackage{amsfonts}       % blackboard math symbols
\usepackage{nicefrac}       % compact symbols for 1/2, etc.
\usepackage{microtype}      % microtypography
\usepackage{subfigure}
\usepackage{multirow}
\usepackage{multicol}
\usepackage{stfloats}
\usepackage{caption}
\usepackage{graphicx}
\usepackage{xcolor}
\usepackage{amsmath}
\usepackage{bm}
\usepackage{enumitem}

\title{Towards Deeper Graph Neural Networks with Differentiable Group Normalization}

\author{%
  Kaixiong Zhou \\
  Texas A\&M University\\
  \texttt{zkxiong@tamu.edu} \\
   \And
   Xiao Huang \\
   The Hong Kong Polytechnic University \\
   \texttt{xhuang.polyu@gmail.com} \\
   \And
   Yuening Li \\
   Texas A\&M University \\
   \texttt{liyuening@tamu.edu} \\
   \And
   Daochen Zha \\
   Texas A\&M University \\
   \texttt{daochen.zha@tamu.edu} \\
   \And
   Rui Chen \\
   Samsung Research America \\
   \texttt{rui.chen1@samsung.com} \\
   \And
   Xia Hu \\
   Texas A\&M University \\
   \texttt{xiahu@tamu.edu}
}

\begin{document}

\maketitle

\begin{abstract}
% Graph neural networks (GNNs) have achieved prominent performance in network learning tasks. They learn the representation of a node by aggregating its neighbors. This key mechanism, as the number of layers increases, would cause an over-smoothing issue and limit the performance of GNNs. Existing studies focus on analyzing and improving the node pair distance. However, their solutions tend to achieve sub-optimal results since the intrinsic community structures have been ignored. The representations of nodes within the same community/class need be similar to facilitate a downstream classifier, while the different classes are expected to be separated in embedding space. To bridge the gap, we introduce two over-smoothing metrics and a novel technique - differentiable group normalization (DGN). It normalizes nodes within the same group independently to increase their smoothness, and separates node distributions among different groups to significantly alleviate the over-smoothing issue. Experiments on real-world datasets demonstrate that DGN makes GNN models more robust to over-smoothing and achieves better performance with deeper GNNs.

%\textcolor{red}{and existing studies mainly focus on analyzing and improving the node pair distance.} 

Graph neural networks (GNNs), which learn the representation of a node by aggregating its neighbors, have become an effective computational tool in downstream applications. Over-smoothing is one of the key issues which limit the performance of GNNs as the number of layers increases. It is because the stacked aggregators would make node representations converge to indistinguishable vectors. Several attempts have been made to tackle the issue by bringing linked node pairs close and unlinked pairs distinct. However, they often ignore the intrinsic community structures and would result in sub-optimal performance. The representations of nodes within the same community/class need be similar to facilitate the classification, while different classes are expected to be separated in embedding space. To bridge the gap, we introduce two over-smoothing metrics and a novel technique, i.e., differentiable group normalization (DGN). It normalizes nodes within the same group independently to increase their smoothness, and separates node distributions among different groups to significantly alleviate the over-smoothing issue. Experiments on real-world datasets demonstrate that DGN makes GNN models more robust to over-smoothing and achieves better performance with deeper GNNs.

\end{abstract}

\section{Introduction}
Graph neural networks (GNNs)~\cite{scarselli2008graph, li2015gated, wu2019comprehensive} have emerged as a promising tool for analyzing networked data, such as biochemical networks~\cite{duvenaud2015convolutional, xu2018powerful}, social networks~\cite{hamilton2017inductive, huang2019graph}, and academic networks~\cite{gao2018large, zhou2019auto}. The successful outcomes have led to the development of many advanced GNNs, including graph convolutional networks~\cite{kipf2016semi}, graph attention networks~\cite{velickovic2017graph}, and simple graph convolution networks~\cite{wu2019simplifying}.

%Following the message passing strategy~\cite{hamilton2017inductive}, GNN learn a node's embedding representations via aggregating representations of its neighbors and itself. The learned node representations could be applied for a variety of tasks on graphs effectively, such as node classification and link prediction. Accompanied with exploration of GNN variants and their breakthrough results, it is also crucial to understand the working mechanisms and limitations of GNN. Such analysis could help us develop the advanced GNN models to fully exploit the powers of deep learning models in non-Euclidian graph data. 

% mixing the features of a node with its neighbors based on the mean or sum aggregator iteratively, 
Besides the exploration of graph neural network variants in different applications, understanding the mechanism and limitation of GNNs is also a crucial task. The core component of GNNs, i.e., a neighborhood aggregator updating the representation of a node iteratively via mixing itself with its neighbors' representations~\cite{hamilton2017inductive, zhou2019multi}, is essentially a low-pass smoothing operation~\cite{nt2019revisiting}. It is in line with graph structures since the linked nodes tend to be similar~\cite{mcpherson2001birds}.
It has been reported that, as the number of graph convolutional layers increases, all node representations over a graph will converge to indistinguishable vectors, and GNNs perform poorly in downstream applications~\cite{li2018deeper, oono2020graph,li2019specae}. It is recognized as an over-smoothing issue. Such an issue prevents GNN models from going deeper to exploit the multi-hop neighborhood structures and learn better node representations.

%As widely studied in multiple recent papers, the key working mechanism and limitation with GNNs are their smoothing and over-smoothing, respectively. On the one hand, the main reason GNNs outperform existing Euclidean-based methods is their message passing strategy over the graph structure. GNNs mix the features of a node and its neighbors based on the mean or sum aggregator iteratively, which is essentially a smoothing operation. The resulting similar embeddings greatly ease the downstream learning task, like node classification, since the directly-connected nodes usually have the same label. On the other hand, by stacking multiple layers of feature mixing, all the nodes' embeddings will converge to a stationary point that is independent of weights and inputs. This phenomenon is recognized as the over-smoothing issue exiting in the deep GNN models, which has be studied theoretically and empirically before. The over-smoothing issue hurts classification performance by leading the nodes' representations a few hops away to be indistinguishable, since they belong to different classes possibly. Most of the previous studies focused on analyzing the embedding distances among every linked/distant nodes in the Euclidian space. Some techniques are then proposed to keep the embedding distance of distant nodes to relieve the over-smoothing issues, including  distance regularization, skip connection and edge dropping. 

A lot of efforts have been devoted to alleviating the over-smoothing issue, such as regularizing the node distance~\cite{chen2019measuring}, node/edge dropping~\cite{rong2020dropedge, hou2020measuring}, batch and pair normalizations~\cite{dwivedi2020benchmarking, ioffe2015batch, zhao2019pairnorm}. Most of existing studies focused on measuring the over-smoothing based on node pair distances. By using these measurements, representations of linked nodes are forced to be close to each other, while unlinked pairs are separated.
Unfortunately, the global graph structures and group/community characteristics are ignored, which leads to sub-optimal performance. For example, to perform node classification, an ideal solution is to assign similar vectors to nodes in the same class, instead of only the connected nodes. In the citation network Pubmed~\cite{yang2016revisiting}, $36\%$ of unconnected node pairs belong to the same class. These node pairs should instead have a small distance to facilitate node classification. Thus, we are motivated to tackle the over-smoothing issue in GNNs from a group perspective.

Given the complicated group structures and characteristics, it remains a challenging task to tackle the over-smoothing issue in GNNs.
First, the formation of over-smoothing is complex and related to both local node relations and global graph structures, which makes it hard to measure and quantify.
Second, the group information is often not directly available in real-world networks. This prevents existing tools such as group normalization being directly applied to solve our problem~\cite{wu2018group}. For example, while the group of adjacent channels with similar features could be directly accessed in convolutional neural networks~\cite{krizhevsky2012imagenet}, it is nontrivial to cluster a network in a suitable way. The node clustering needs to be in line with the embeddings and labels, during the dynamic learning process.

%Second, an efficient solution is required to optimize the structure based metrics to improve node representation learning.
%Second, the available tools, such as batch normalization and group normalization, can not effectively solve the over-smoothing issue without the attention on graph structures. For example, the pair normalization is designed to simply improve the average node pair distance no matter the nodes locates in same community or not~\cite{zhao2019pairnorm}. 
% Third, batch/group normalization can not effectively solve the over-smoothing issue in GNNs. 
To bridge the gap, in this paper, we perform a quantitative study on the over-smoothing in GNNs from a group perspective. We aim to answer two research questions. First, how can we precisely measure the over-smoothing in GNNs? Second, how can we handle over-smoothing in GNNs? Through exploring these questions, we make three significant contributions as follows. 
\begin{itemize}
%[wide=0pt, leftmargin=\dimexpr\labelwidth + 2\labelsep\relax]
    \item Present two metrics to quantify the over-smoothing in GNNs: (1) Group distance ratio, clustering the network and measuring the ratio of inter-group representation distance over intra-group one; (2) Instance information gain, treating node instance independently and measuring the input information loss during the low-pass smoothing. 
    %We design two metrics to understand the over-smoothing issue in GNN from the perspective of node groups. 
    \item Propose differentiable group normalization to significantly alleviate over-smoothing. It softly clusters nodes and normalizes each group independently, which prevents distinct groups from having close node representations to improve the over-smoothing metrics. 
    %It enables better performance with deeper GNNs.
\item Empirically show that deeper GNNs, when equipped with the proposed differentiable group normalization technique, yield better node classification accuracy.
\end{itemize}

\section{Quantitative Analysis of Over-smoothing Issue}
In this work, we use the semi-supervised node classification task as an example and illustrate how to handle the over-smoothing issue.
A graph is represented by $G = \{\mathcal{V}, \mathcal{E}\}$, where $\mathcal{V}$ and $\mathcal{E}$ represent the sets of nodes and edges, respectively. Each node $v \in \mathcal{V}$ is associated with a feature vector $x_v \in \mathbb{R}^{d}$ and a class label $y_v$. Given a training set $\mathcal{V}_l$ accompanied with labels, the goal is to classify the nodes in the unlabeled set $\mathcal{V}_u = \mathcal{V}\setminus \mathcal{V}_l$ via learning the mapping function based on GNNs.

% The goal of node classification task is to learn a representation vector $h_v$ and a mapping function $f(h_v)$ to predict the class label $y_v$ of node $v$. 

\subsection{Preliminaries} 
% We build upon GNNs to learn representations vector $h_v$ in an end-to-end fashion. 
Following the message passing strategy~\cite{gilmer2017neural}, GNNs update the representation of each node via aggregating itself and its neighbors' representations. Mathematically, at the $k$-th layer, we have,
\begin{equation}
    \label{eq:GNN}
        N^{(k)}_v = \mathrm{AGG}(\{a^{(k)}_{vv'} W^{(k)} h^{(k-1)}_{v'}: v'\in \mathcal{N}(v)\}), \quad
        h^{(k)}_v = \mathrm{COM}(a^{(k)}_{vv}W^{(k)}h^{(k-1)}_v, N^{(k)}_v).
\end{equation}
$N^{(k)}_v$ and $h^{(k)}_v$ denote the aggregated neighbor embedding and embedding of node $v$, respectively. We initialize $h^{(0)}_v = x_v$. $\mathcal{N}(v) = \{v' | e_{v, v'} \in \mathcal{E}\}$ represents the set of neighbors for node $v$, where $e_{v, v'}$ denotes the edge that connects nodes $v$ and $v'$. $W^{(k)}$ denotes the trainable matrix used to transform the embedding dimension. $a^{(k)}_{vv'}$ is the link weight over edge $e_{v, v'}$, which could be determined based on the graph topology or learned by an attention layer. Symbol $\mathrm{AGG}$ denotes the neighborhood aggregator usually implemented by 
a summation 
pooling. To update node $v$, function $\mathrm{COM}$ is applied to combine neighbor information and node embedding from the previous layer. It is observed that the weighted average in Eq.~\eqref{eq:GNN} smooths node embedding with its neighbors to make them similar. For a full GNN model with $K$ layers, the final node representation is given by $h_v = h^{(K)}_v$, which captures the neighborhood structure information within $K$ hops.

\subsection{Measuring Over-smoothing with Group Structures}
In GNNs, the neighborhood aggregation strategy smooths nodes' representations over a graph~\cite{nt2019revisiting}. It will make the representations of nodes converge to similar vectors as the number of layers $K$ increases. This is called the over-smoothing issue, and would cause the performance of GNNs deteriorates as $K$ increases. To address the issue, the first step is to measure and quantify the over-smoothing~\cite{chen2019measuring, hou2020measuring}. Measurements in existing work are mainly based on the distances between node pairs~\cite{rong2020dropedge, zhao2019pairnorm}.
A small distance means that a pair of nodes generally have undistinguished representation vectors, which might triggers the over-smoothing issue. 

However, the over-smoothing is also highly related to global graph structures, which have not been taken into consideration.
For some unlinked node pairs, we would need their representations to be close if they locate in the same class/community, to facilitate the node classification task. Without the specific group information, the metrics based on pair distances may fail to indicate the over-smoothing.
Thus, we propose two novel over-smoothing metrics, i.e., group distance ratio and instance information gain. They quantify the over-smoothing from global (communities/classes/groups) and local (node individuals) views, respectively.

\bm{\mathrm{Definition 1 (Group Distance Ratio).}}
Suppose that there are $C$ classes of node labels. We intuitively cluster nodes of the same class label into a group to formulate the labeled node community. Formally, let $\bm{L}_i = \{h_{iv}\}$ denote the group of representation vectors, where node $v$ is associated with label $i$. We have a series of labeled groups $\{\bm{L}_1, \cdots, \bm{L}_C\}$. Group distance ratio $R_{\mathrm{Group}}$ measures the ratio of inter-group distance over intra-group distance in the Euclidean space. We have:
\begin{equation}
    \label{equ: GDR}
    R_{\mathrm{Group}} = \frac{\frac{1}{(C-1)^2}\sum_{i\neq j}(\frac{1}{|\bm{L}_i||\bm{L}_j|}\sum_{h_{iv}\in \bm{L}_i}\sum_{h_{jv'}\in \bm{L}_j} || h_{iv} - h_{jv'} ||_2)} {\frac{1}{C}\sum_{i}(\frac{1}{|\bm{L}_i|^2}\sum_{h_{iv}, h_{iv'}\in \bm{L}_i} || h_{iv} - h_{iv'} ||_2)},
\end{equation}
where $||\cdot||_2$ denotes the L2 norm of a vector and $|\cdot|$ denotes the set cardinality. The numerator (denominator) represents the average of pairwise representation distances between two different groups (within a group). One would prefer to reduce the intra-group distance to make representations of the same class similar, and increase the inter-group distance to relieve the over-smoothing issue. On the contrary, a small $R_{\mathrm{Group}}$ leads to the over-smoothing issue where all groups are mixed together, and the intra-group distance is maintained to hinder node classification.
%It is easy to be observed that metric $R_{\mathrm{Group}}$ is positively related to $S_{\mathrm{Group}}$. We give metric $R_{\mathrm{Group}}$ to analyze over-smoothing issue more intuitively in the Eucilidian space from group perspective.  

\bm{\mathrm{Definition 2 (Instance Information Gain).}} In an attributed network, a node's feature decides its class label to some extent. We treat each node instance independently, and define instance information gain $G_{\mathrm{Ins}}$ as how much input feature information is contained in the final representation. Let $\mathcal{X}$ and $\mathcal{H}$ denote the random variables of input feature and representation vector, respectively. We define their probability distributions with $P_{\mathcal{X}}$ and  $P_{\mathcal{H}}$, and use $P_{\mathcal{XH}}$ to denote their joint distribution. $G_{\mathrm{Ins}}$ measures the dependency between node feature and representation via their mutual information:
\begin{equation}
    \label{equ: IIS}
    G_{\mathrm{Ins}} = I(\mathcal{X}; \mathcal{H}) = \sum_{x_v\in \mathcal{X}, h_v \in \mathcal{H}} P_{\mathcal{XH}}(x_v, h_v) \log \frac{P_{\mathcal{XH}}(x_v, h_v)}{P_{\mathcal{X}}(x_v)P_{\mathcal{H}}(h_v)}.
\end{equation}
% \bm{\mathrm{Definition 2 (Instance Information Gain).}} Let $\mathcal{X}$ denote the random variable of input feature, whose probability density distribution $P_{\mathcal{X}}$ is estimated by nonparametric methods with a set of sampled features $\{x_1, \cdots, x_n\}$. Each node feature is sampled with probability $1 / |\mathcal{V}|$ following the empirical uniform distribution, where $|\mathcal{V}|$ denote the number of nodes in a graph. Similarly, we define $\mathcal{H}$ as the random variable of representation vector, whose probability density distribution is denoted as $P_{\mathcal{H}}$. Let $P_{\mathcal{XH}}$ denote the joint distribution of variables $X$ and $H$. Instance information smoothness $G_{\mathrm{Ins}}$ measures how much of input feature is contained in the learned representation by their mutual information. Formally,
% \begin{equation}
%     \label{equ: IIS}
%     G_{\mathrm{Ins}} = I(\mathcal{X}; \mathcal{H}) = \sum_{x_v\in \mathcal{X}, h_v \in \mathcal{H}} P_{\mathcal{XH}}(x_v, h_v) \log \frac{P_{\mathcal{XH}}(x_v, h_v)}{P_{\mathcal{X}}(x_v)P_{\mathcal{H}}(h_v)} = H({\mathcal{X}}) - H({\mathcal{X}} | {\mathcal{H}}).
% \end{equation}
We list the details of variable definitions and mutual information calculation in the context of GNNs in Appendix. With the intensification of the over-smoothing issue, nodes average the neighborhood information and lose their self features, which leads to a small value of $G_{\mathrm{Ins}}$. 

\subsection{Illustration of Proposed Over-smoothing Metrics}
% We analyze the over-smoothing issue based on the two proposed metric: $G_{\mathrm{Ins}}$ and $S_{\mathrm{Group}}$. Apart from the over-smoothing issues, the increasing parameters as layer number adding also lead to overfitting problem and vanishing gradient in backpropagation training. In order to focus on studying the over-smoothing issue, we apply the simplified GNN model, SGC, which removes all trainable projection matrices and all nonlinear activations between layers. The model details are introduced in Appendix. 
% Formally, the class prediction of SGC is written as $\hat{Y} = \mathrm{softmax}(S^KXW) \in \mathbb{R}^{n\times C}$. $X$ and $\hat{Y}$ denote feature matrix and predicted probability matrix of all nodes in a graph, respectively. $S \in \mathbb{R}^{n\times n}$ denotes the normalized adjacency matrix with added self-loops, and $W \in \mathbb{R}^{n\times C}$ denote trainable parameter of classifier. It has been shown that SGC does not greatly sacrifice the classification accuracy accompanied with the model simplification. 

\begin{figure}
    \centering
    \includegraphics[width=1\textwidth]{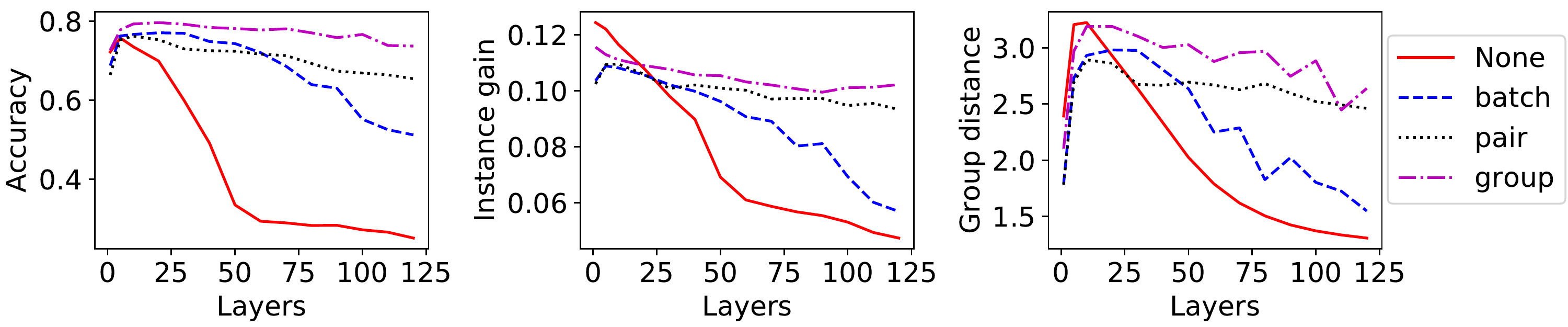}
    \caption{The test accuracy, instance information gain, and group distance ratio of SGC on Cora. We compare differentiable group normalization with none, batch and pair normalizations.}
    \label{fig:SGC_Cora}
\end{figure}

Based on the two proposed metrics, we take simple graph convolution networks (SGC) as an example, and analyze the over-smoothing issue on Cora dataset~\cite{yang2016revisiting}. SGC simplifies the model through removing all the trainable weights between layers to avoid the potential of overfitting~\cite{wu2019simplifying}. So the over-smoothing issue would be the major cause of performance dropping in SGC. As shown by the red lines in Figure \ref{fig:SGC_Cora}, the graph convolutions first exploit neighborhood information to improve test accuracy up to $K = 5$, after which the over-smoothing issue starts to worsen the performance. At the same time, instance information gain $G_{\mathrm{Ins}}$ and group distance ratio $R_{\mathrm{Group}}$ decrease due to the over-smoothing issue. For the extreme case of $K = 120$, the input features are filtered out and all groups of nodes converge to the same representation vector, leading to $G_{\mathrm{Ins}} = 0$ and  $R_{\mathrm{Group}} = 1$, respectively. Our metrics quantify the smoothness of node representations based on group structures, but also have the similar variation tendency with test accuracy to indicate it well.

% With the increment of layer $K$, red lines in Figure \ref{fig:SGC_Cora} illustrates the proposed metrics and test accuracy of SGC on dataset Cora. We can observe that smoothing graph convolutions first improve test accuracy up to $K = 5$, after which over-smoothing issue starts worsening the performance. Notably, smoothing graph convolutions update a node's representation vector with a weighted average of itself and its neighbors. As layer $K$ increases, the input information retained in representation keeps decreasing as shown by instance gain $G_{\mathrm{Ins}}$. At the same time, group distance $D_{\mathrm{Group}}$ increases at the early stage and then decreases. Since the densely-connected nodes usually share the same class label, a small number of smoothing layers aggregates helpful information to improve intra-group representation similarity and inter-group representation distance. However, over-smoothing issue appears at the larger numbers of $K$, at which nodes' representations in a graph tends to be similar and group distance $D_{\mathrm{Group}}$ drops as a result. At the extreme case of $K = 120$, all of nodes' representations converge to the same value and become totally indistinguishable, leading to the limit group distance of $D_{\mathrm{Group}} = 1$.

\section{Differentiable Group Normalization}
% To better illustrate the research purpose of this work,
We start with a graph-regularized optimization problem~\cite{kipf2016semi, chen2019measuring}. To optimize the over-smoothing metrics of $G_{\mathrm{Ins}}$ and $R_{\mathrm{Group}}$, one traditional approach is to minimize the loss function:
\begin{equation}
\label{equ: objective}
\mathcal{L} = \mathcal{L}_0 - G_{\mathrm{Ins}} - \lambda R_{\mathrm{Group}}.
\end{equation}
$\mathcal{L}_0$ denotes the supervised cross-entropy loss w.r.t. representation probability vectors $h_v\in \mathbb{R}^{C\times 1}$ and class labels. $\lambda$ is a balancing factor. The goal of optimization problem Eq.~\eqref{equ: objective} is to learn node representations close to the input features and informative for their class labels. Considering the labeled graph communities, it also improves the intra-group similarity and inter-group distance. However, it is non-trivial to optimize this objective function due to the non-derivative of non-parametric statistic $G_{\mathrm{Ins}}$~\cite{kolchinsky2017estimating, kolchinsky2019nonlinear} and the expensive computation of $R_{\mathrm{Group}}$.

\subsection{Proposed Technique for Addressing Over-smoothing}

Instead of directly optimizing regularized problem in Eq.~\eqref{equ: objective}, we propose the differentiable group normalization (DGN) applied between graph convolutional layers to normalize the node embeddings group by group. The key intuition is to cluster nodes into multiple groups and then normalize them independently. Consider the labeled node groups (or communities) in networked data. The node embeddings within each group are expected to be rescaled with a specific mean and variance to make them similar. Meanwhile, the embedding distributions from different groups are separated by adjusting their means and variances. We develop an analogue with the group normalization in convolutional neural networks (CNNs)~\cite{wu2018group}, which clusters a set of adjacent channels with similar characteristics into a group and treats it independently. Compared with standard CNNs, the challenge in designing DGN is how to cluster nodes in a suitable way. The clustering needs to be in line with the embedding and labels, during the dynamic learning process.
%There are no natural notions of labeled node group, i.e., one cannot simply cluster a patch of linked nodes which may have different class labels. 
%Furthermore, the node labels in the test set cannot be accessed during training and model inference.  

We address this challenge by learning a cluster assignment matrix, which softly maps nodes with close embeddings into a group. Under the supervision of training labels, the nodes close in the embedding space tend to share a common label. To be specific, we first describe how DGN clusters and normalizes nodes in a group-wise fashion given an assignment matrix. After that, we discuss how to learn the assignment matrix to support differentiable node clustering.

\noindent{\bm{\mathrm{Group Normalization.}}} Let $H^{(k)} = [h_1^{(k)} ,\cdots, h_n^{(k)}]^T \in \mathbb{R}^{n\times d^{(k)}}$ denote the embedding matrix generated from the $k$-th graph convolutional layer. Taking $H^{(k)}$ as input, DGN softly assigns nodes into groups and normalizes them independently to output a new embedding matrix for the next layer. Formally, we define the number of groups as $G$, and denote the cluster assignment matrix by $S^{(k)} \in \mathbb{R}^{n\times G}$. $G$ is a hyperparameter that could be tuned per dataset. The $i$-th column of $S^{(k)}$, i.e., $S^{(k)}[:, i]$, indicates the assignment probabilities of nodes in a graph to the $i$-th group.
% Each row of $S^{(k)}$ corresponds to the probability vector of a node belonging to all the $G$ groups. %Intuitively, $S^{(k)}$ provides a soft assignment of nodes over a graph into $G$ groups.
Supposing that $S^{(k)}$ has already been computed, we cluster and normalize nodes in each group as follows:
\begin{equation}
    \label{equ:cluster_norm}
    H^{(k)}_i = S^{(k)}[:, i] \circ H^{(k)} \in \mathbb{R}^{n\times d^{(k)}};\quad \tilde H^{(k)}_i = \gamma_i (\frac{H^{(k)}_i - \mu_i}{\sigma_i}) + \beta_i \in \mathbb{R}^{n\times d^{(k)}}.
\end{equation}
Symbol $\circ$ denotes the row-wise multiplication. The left part in the above equation represents the soft node clustering for group $i$, whose embedding matrix is given by $H^{(k)}_i$. The right part performs the standard normalization operation. In particular, $\mu_i$ and $\sigma_i$ denote the vectors of running mean and standard deviation of group $i$, respectively, and $\gamma_i$ and $\beta_i$ denote the trainable scale and shift vectors, respectively. Given the input embedding $H^{(k)}$ and the series of normalized embeddings $\{\tilde H^{(k)}_1, \cdots, \tilde H^{(k)}_G\}$, DGN % combines them linearly to 
generates the final embedding matrix $\tilde{H}^{(k)}$ for the next layer as follows:
\begin{equation}
    \label{equ: final}
    \tilde{H}^{(k)} =  H^{(k)} + \lambda \sum_{i=1}^{G} \tilde H^{(k)}_i \in \mathbb{R}^{n\times d^{(k)}}.
\end{equation}
$\lambda$ is a balancing factor as mentioned before. Inspecting the loss function in Eq.~\eqref{equ: objective}, DGN utilizes components $H^{(k)}$ and $\sum_{i=1}^{G} \tilde H^{(k)}_i$ to improve terms $G_{\mathrm{Ins}}$ and $R_{\mathrm{Group}}$, respectively. In particular, we preserve the input embedding $H^{(k)}$ to avoid over-normalization and keep the input feature of each node to some extent. Note that the linear combination of $H^{(k)}$ in DGN is different from the skip connection in GNN models~\cite{li2019deepgcns, gurel2019anatomy}, which instead connects the embedding output $H^{(k-1)}$ from the last layer. The technique of skip connection could be included to further boost the model performance. Group normalization $\sum_{i=1}^{G} \tilde H^{(k)}_i$ rescales the node embeddings within each group independently to make them similar. Ideally, we assign the close node embeddings with a common label to a group. Node embeddings of the group are then distributed closely around the corresponding running mean. Thus for different groups associate with distinct node labels, we disentangle their running means and separates the node embedding distributions.
%and separates running means as well as embedding distributions among different groups by normalizing them independently. 
By applying DGN between the successive graph convolutional layers, we are able to optimize Problem (\ref{equ: objective}) to 
%learn node representations close to their input features, and at the same time 
mitigate the over-smoothing issue. 

\noindent{\bm{\mathrm{Differentiable Clustering.}}} We apply a linear model to compute the cluster assignment matrix $S^{(k)}$ used in Eq.~\eqref{equ:cluster_norm}. The mathematical expression is given by:
\begin{equation}
    \label{equ: cluster_matrix}
    S^{(k)} = \mathrm{softmax}(H^{(k)} U^{(k)}).
\end{equation}
$U^{(k)} \in \mathbb{R}^{d^{(k)}\times G}$ denotes the trainable weights for a DGN module applied after the $k$-th graph convolutional layer. $\mathrm{softmax}$ function is applied in a row-wise way to produce the normalized probability vector w.r.t all the $G$ groups for each node. Through the inner product between $H^{(k)}$ and $U^{(k)}$, the nodes with close embeddings are assigned to the same group with a high probability. Here we give a simple and effective way to compute $S^{(k)}$. Advanced neural networks could be applied.
%to improve node clustering in the future. 

% In the following we describe how DGN generates cluster assignment matrix $S^{(k)}$ used in Eq.~\eqref{equ:cluster_norm}. An intuitive solution is to assign nodes into their nearby groups with high probabilities base on the Eucilidian distance between embedding $H^{(k)}$ and running mean $\bar\mu_i \in \mathbb{R}^{d^{(k)}\times 1}$ at each group. Let $\bar{U} = [\bar\mu_1, \cdots, \bar\mu_G]^T \in \mathbb{R}^{G\times d^{(k)}}$ denote the running mean matrix of all groups. We generate matrix $S^{(k)}$ through measuring the dot product distance for each node-group pair as follows:
% \begin{equation}
%     \label{equ: cluster_matrix}
%     S^{(k)} = \mathrm{softmax}(H^{(k)} \bar{U}^T),
% \end{equation}

% \noindent{\bm{\mathrm{Auxiliary Distribution Regularization.}}} To clearly separate the embedding distributions of different groups, one can further regularize loss function in Eq.~\eqref{equ: objective} by embedding distribution distance $\mathcal{L}_{\mathrm{Dis}} = \sum_{i}^G ||\gamma_i||_2 -\sum_{i\neq j}||\beta_i - \beta_j||_2$. The first term measures the standard deviations of all groups, which should be small to avoid the overlap among groups. The other term quantifies the center distance between every two groups, which should be maximized to separate different groups. In practice, we find that adding such regularization could help improve classification performance on some datasets.

\noindent{\bm{\mathrm{Time Complexity Analysis.}}} Suppose that the time complexity of embedding normalization at each group is $\mathcal{O}(T)$, where $T$ is a constant depending on embedding dimension $d^{(k)}$ and node number $n$. The time cost of group normalization $\sum_{i=1}^{G} \tilde H^{(k)}_i$ is  $\mathcal{O}(GT)$. Both the differentiable clustering (in Eq.~\eqref{equ:cluster_norm}) and the linear model (in Eq.~\eqref{equ: cluster_matrix}) have a time cost of $\mathcal{O}(nd^{(k)}G)$. Thus the total time complexity of a DGN layer is given by $\mathcal{O}(nd^{(k)}G + GT)$, which linearly increases with $G$.

\noindent{\bm{\mathrm{Comparison with Prior Work.}}} To the best of our knowledge, the existing work mainly focuses on analyzing and improving the node pair distance to relieve the over-smoothing issue~\cite{chen2019measuring, hou2020measuring, zhao2019pairnorm}. One of the general solutions is to train GNN models regularized by the pair distance~\cite{chen2019measuring}. Recently, there are two related studies applying batch normalization~\cite{dwivedi2020benchmarking} or pair normalization~\cite{zhao2019pairnorm} to keep the overall pair distance in a graph. Pair normalization is a ``slim'' realization of batch normalization by removing the trainable scale and shift. However, the metric of pair distance and the resulting techniques ignore global graph structure, and may achieve sub-optimal performance in practice. In this work, we measure over-smoothing of GNN models based on communities/groups and independent node instances. We then formulate the problem in Eq.~\eqref{equ: objective} to optimize the proposed metrics, and propose DGN to solve it in an efficient way, which in turn addresses the over-smoothing issue.

%The technical details of BN and PN are listed in Appendix.  
% Compared with other normalization methods, DGN shows its effectiveness and robustness in exploring the deep GNN models. 
\subsection{Evaluating Differentiable Group Normalization on Attributed Graphs}
We apply DGN to the SGC model to validate its effectiveness in relieving the over-smoothing issue. Furthermore, we compare with the other two available normalization techniques used upon GNNs, i.e., batch normalization and pair normalization.
As shown in Figure~\ref{fig:SGC_Cora}, the test accuracy of DGN remains stable with the increase in the number of layers. By preserving the input embedding and normalizing node groups independently, DGN achieves superior performance in terms of instance information gain as well as group distance ratio. 
The promising results indicate that our DGN tackles the over-smoothing issue more effectively, compared with none, batch and pair normalizations.
%This outperforming result is credited with the preserving of original embedding and the independent group normalization in Eq.~\eqref{equ: final}. In particular, embedding $H^{(k)}$ keeps the original input feature informative for node's class label, which is revealed by the slower decay of instance gain $G_{\mathrm{Ins}}$ at the middle of Figure~\ref{fig:SGC_Cora}. $\sum_{i=1}^{G} \tilde H^{(k)}_i$ normalizes each group independently to improve intra-group similarity and inter-group distance, which is validated by group distance $D_{\mathrm{Group}}$ at the right of Figure~\ref{fig:SGC_Cora}. 

It should be noted that, the highest accuracy of $79.7\%$ is achieved with DGN when $K = 20$. This observation  contradicts with the common belief that GNN models work best with a few layers on current benchmark datasets~\cite{zhou2018graph}. With the integration of advanced techniques, such as DGN, we are able to exploit deeper GNN architectures to unleash the power of deep learning in network analysis. 

\subsection{Evaluation in Scenario with Missing Features}
To further illustrate that DGN could enable us to achieve better performance with deeper GNN architectures, we apply it to a more complex scenario. We assume that the attributes of nodes in the test set are missing. It is a common scenario in practice~\cite{zhao2019pairnorm}. For example, in social networks, new users are often lack of profiles and tags~\cite{rashid2008learning}. To perform prediction tasks on new users, we would rely on the node attributes of existing users and their connections to new users. 
In such a scenario, we would like to apply more layers to exploit the neighborhood structure many hops away to improve node representation learning. Since the over-smoothing issue gets worse with the increasing of layer numbers, the benefit of applying normalization will be more obvious in this scenario.

\begin{figure}[t]
    \centering
    \includegraphics[width=1\textwidth]{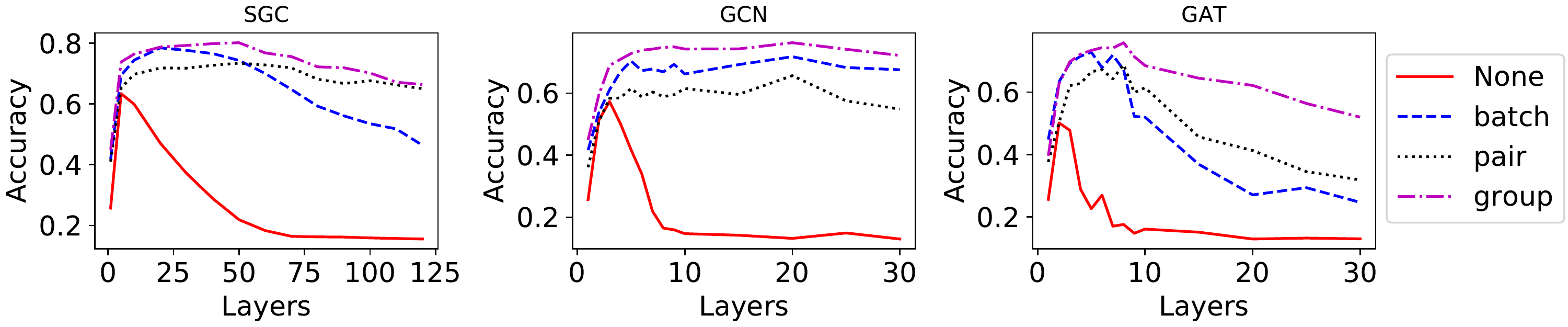}
    \caption{The test accuracies of SGC, GCN, and GAT models on Cora with missing features. We compare differentiable group normalization with none, batch and pair normalizations.}
    \label{fig:all_Cora}
\end{figure}

%to show the generalization capability of DGN in enabling deeper GNNs
We remove the input features of both validation and test sets in Cora, and replace them with zeros~\cite{zhao2019pairnorm}. Figure~\ref{fig:all_Cora} presents the results on three widely-used models, i.e., SGC, graph convolutional networks (GCN), and graph attention networks (GAT). Due to the over-smoothing issue, GNN models without any normalization fail to distinguish nodes quickly with the increasing number of layers. In contrast, the normalization techniques reach their highest performance at larger layer numbers, after which they drop slowly. We observe that DGN obtains the best performance with $50$, $20$, and $8$ layers for SGC, GCN, and GAT, respectively. These layer numbers are significantly larger than those of the widely-used shallow models (e.g., two or three layers).

% Compared batch and pair normalizations, DGN obtains the average improvements of {\color{red}in terms of the highest test accuracy}.  

\section{Experiments}
We now empirically evaluate the effectiveness and robustness of DGN on real-world datasets. We aim to answer three questions as follows. \textbf{Q1:} Compared with the state-of-the-art normalization methods, can DGN alleviate the over-smoothing issue in GNNs in a better way? {\bf Q2:} Can DGN help GNN models achieve better performance by enabling deeper GNNs?
%How does DGN compare to other normalization methods on both attribute and plain graph? 
\textbf{Q3:} How do the hyperparameters influence the performance of DGN? %\textbf{Q3:} How does DCN softly clusters nodes into meaningful groups to separate their representations?
%\begin{itemize}[wide=0pt, leftmargin=\dimexpr\labelwidth + 2\labelsep\relax]
%    \item  
%\end{itemize}

\subsection{Experiment Setup}

% \paragraph{Datasets.}
\textbf{Datasets.}
Joining the practice of previous work, we evaluate GNN models by performing the node classification task on four datasets: Cora, Citeseer, Pubmed~\cite{yang2016revisiting}, and CoauthorCS~\cite{shchur2018pitfalls}. We also create graphs by removing features in validation and test sets. The dataset statistics are in Appendix. % {\red why we use accuracy, not F1 score?}

% \paragraph{Implementations.} 
\textbf{Implementations.} 
Following the previous settings, we choose the hyperparameters of GNN models and optimizer as follows. We set the number of hidden units to $16$ for GCN and GAT models. The number of attention heads in GAT is $1$. Since a larger parameter size in GCN and GAT may lead to overfitting and affects the study of over-smoothing issue, we compare normalization methods by varying the number of layers $K$ in $\{1, 2, \cdots, 10, 15, \cdots, 30\}$. For SGC, we increase the testing range and vary $K$ in $\{1, 5, 10, 20, \cdots, 120\}$. We train with a maximum of $1000$ epochs using the Adam optimizer~\cite{kingma2014adam} and early stopping. Weights in GNN models are initialized with Glorot algorithm~\cite{glorot2010understanding}. We use the following sets of hyperparameters for Citeseer, Cora, CoauthorCS: $0.6$ (dropout rate), $5\cdot 10^{-4}$ (L2 regularization), $5\cdot 10^{-3}$ (learning rate), and for Pubmed: $0.6$ (dropout rate), $1\cdot 10^{-3}$ (L2 regularization), $1\cdot 10^{-2}$ (learning rate). We
run each experiment $5$ times and report the average.

% \paragraph{Baselines.} 
\textbf{Baselines.} 
We compare with none normalization (NN), batch normalization (BN)~\cite{dwivedi2020benchmarking, ioffe2015batch} and pair normalization (PN)~\cite{zhao2019pairnorm}. Their technical details are listed in Appendix.

% \paragraph{DGN Configurations.} 
\textbf{DGN Configurations.} 
The key hyperparameters include group number $G$ and balancing factor $\lambda$. 
%Note that we focus on comparison between normalization methods, rather than achieving the state of the art performance of node classification on these benchmark datasets. We choose default parameters by roughly trying a few combinations of $G$ and $\lambda$. 
Depending on the number of class labels, we apply $5$ groups to Pubmed and $10$ groups to the others. The criterion is to use more groups to separate representation distributions in networked data accompanied with more class labels. $\lambda$ is tuned on validation sets to find a good trade-off between preserving input features and group normalization. We introduce the selection of $\lambda$ in Appendix. 

% selected from $\{1\cdot 10^{-3}, 5\cdot 10^{-3}, 1\cdot 10^{-2}, 3\cdot 10^{-2}\}$, based on the criterion of using larger value in more deeper model to attend on representation normalization and relieve over-smoothing issue. The detail configurations of $G$ and $\lambda$ are listed in Appendix for each dataset. 
%The affects of hyperparameters $G$ and $\lambda$ will be carefully studied in the following experiments. 

\subsection{Experiment Results}
% We first compare test accuracies of applying different  normalizations on the original graph, via keeping the input features of nodes. Then we study how normalization techniques explore the deep GNN architectures to learn the plain graph, which is obtained by removing node features in validation and testing sets.

% \paragraph{Studies on alleviating the over-smoothing problem.} 
\textbf{Studies on alleviating the over-smoothing problem.}
To answer {\bf Q1}, Table~\ref{tab: test-compare} summarizes the results of applying different normalization techniques to GNN models on all datasets. We report the performance of GCN and GAT with $2/15/30$ layers, and SGC with $5/60/120$ layers due to space limit. We provide test accuracies, instance information gain and group distance ratio under all depths in Appendix. It can be observed that DGN has significantly alleviated the over-smoothing issue. Given the same layers, DGN almost outperforms all other normalization methods for all cases and greatly slows down the performance dropping. 
%our DGN achieves the similar or even better performances for all GNN models with only $2/5$ layers, at which the over-smoothing issue is not significant. 
It is because the self-preserved component $H^{(k)}$ in Eq.~\eqref{equ: final} keeps the informative input features and avoids over-normalization to distinguish different nodes. This component is especially crucial for models with a few layers since the over-smoothing issue has not appeared. The other group normalization component in Eq.~\eqref{equ: final} processes each group of nodes independently. 
It disentangles the representation similarity between groups, and hence reduces the over-smoothness of nodes over a graph accompanied with graph convolutions.  

\begin{table}[t!]
\setlength{\abovecaptionskip}{0.cm}
\setlength{\belowcaptionskip}{-0.cm}
\setlength{\tabcolsep}{2.8pt}
  \centering
  %The maximum accuracy for each layer is in bold. 
  \caption{Test accuracy in percentage on attributed networks. Layers $a/b$ denote the layer number $a$ in GCN \& GAT and that of $b$ in SGC. $\#K$ denotes the optimal layer numbers where DGN achieves the highest performance.}
  \label{tab: test-compare}
  \begin{tabular}{c|c|cccc|cccc|cccc|c}
    \toprule
    \multirow{2}*{Dataset} & \multirow{2}*{Model} & \multicolumn{4}{c|}{Layers 2/5} & \multicolumn{4}{c|}{Layers 15/60} & \multicolumn{4}{c|}{Layers 30/120} & \multirow{2}*{\#K} \\
    \cline{3-14}
    & & NN & BN & PN & DGN & NN & BN & PN & DGN & NN & BN & PN & DGN & \\
    \hline
    \hline
    \multirow{3}*{Cora} & GCN & $\bm{82.2}$ & $73.9$ & $71.0$ & $82.0$ & $18.1$ & $70.3$ & $67.2$ & $\bm{75.2}$ & $13.1$ & $67.2$ & $64.3$ & $\bm{73.2}$ & $2$ \\
    & GAT & $80.9$ & $77.8$ & $74.4$ & $\bm{81.1}$ & $16.8$ & $33.1$ & $49.6$ & $\bm{71.8}$ & $13.0$ & $25.0$ & $30.2$ & $\bm{51.3}$ & $2$ \\
    & SGC & $75.8$ & $76.3$ & $75.4$ & $\bm{77.9}$ & $29.4$ & $72.1$ & $71.7$ & $\bm{77.8}$ & $25.1$ & $51.2$ & $65.5$ & $\bm{73.7}$ & $20$\\
   
    \hline
    \multirow{3}*{Citeseer} & GCN & $\bm{70.6}$ & $51.3$ & $60.5$ & $69.5$ & $15.2$ & $46.9$ & $46.7$ & $\bm{53.1}$ & $9.4$ & $47.9$ & $47.1$ & $\bm{52.6}$ & $2$ \\
    & GAT & $\bm{70.2}$ & $61.5$ & $62.0$ & $69.3$ & $22.6$ & $28.0$ & $41.4$ & $\bm{52.6}$ & $7.7$  & $21.4$ & $33.3$ & $\bm{45.6}$ & $2$ \\
    & SGC & $\bm{69.6}$ & $58.8$ & $64.8$ & $69.5$ & $\bm{66.3}$ & $50.5$ & $65.0$ & $63.4$ & $60.8$ & $47.3$  & $63.1$ & $\bm{64.7}$ & $30$\\
  
    \hline
    \multirow{3}*{Pubmed} & GCN & $79.3$ & $74.9$ & $71.1$ & $\bm{79.5}$ & $22.5$ & $73.7$ & $70.6$ & $\bm{76.1}$ & $18.0$  & $70.4$ & $70.4$ & $\bm{76.9}$ & $2$ \\
    & GAT & $\bm{77.8}$ & $76.2$ & $72.4$ & $77.5$ & $37.5$ & $56.2$ & $68.8$ & $\bm{75.9}$ & $18.0$ & $46.6$  & $58.2$ & $\bm{73.3}$ & $5$ \\
    & SGC & $71.5$  & $76.5$  & $75.8$ & $\bm{76.8}$ & $34.2$ & $75.2$  & $77.1$  & $\bm{77.4}$ & $23.1$ & $71.6$  & $76.7$ & $\bm{77.1}$ & $10$\\
    \hline
    \multirow{3}*{Coauthors} & GCN & $92.3$ & $86.0$ & $77.8$ & $\bm{92.3}$ & $72.2$ & $78.5$ & $69.5$ & $\bm{83.7}$ & $3.3$ & $\bm{84.7}$ & $64.5$ & $84.4$ & $1$\\
    & GAT & $91.5$ & $89.4$ & $85.9$ & $\bm{91.8}$ & $6.0$ & $77.7$ & $53.1$ & $\bm{84.5}$ & $3.3$ & $16.7$ & $48.1$ & $\bm{75.5}$ & $1$\\
    & SGC & $89.9$ & $88.7$ & $86.0$ & $\bm{90.2}$ & $10.2$ & $59.7$ & $76.4$ & $\bm{81.3}$ & $5.8$ & $30.5$ & $52.6$ & $\bm{60.8}$ & $1$\\
\bottomrule
  \end{tabular}
\end{table}

% \paragraph{Studies on enabling deeper and better GNNs.} 
\textbf{Studies on enabling deeper and better GNNs.}
To answer \textbf{Q2}, we compare all of the concerned normalization methods over GCN, GAT, and SGC in the scenario with missing features. As we have discussed, normalization techniques will show their power in relieving the over-smoothing issue and exploring deeper architectures especially for this scenario.
In Table~\ref{tab: test-compare-missing}, Acc represents the best test accuracy yielded by model equipped with the optimal layer number $\#K$. We can observe that DGN significantly outperforms the other normalization methods on all cases. The average improvements over NN, BN and PN achieved by DGN are $37.8\%$, $7.1\%$ and $12.8\%$, respectively. Compared with vanilla GNN models without any normalization layer, the optimal models accompanied with normalization layers (especially for our DGN) usually possess larger values of $\#K$. It demonstrates that DGN enables to explore deeper architectures to exploit neighborhood information with more hops away by tackling the over-smoothing issue. We present the comprehensive analyses in terms of test accuracy, instance information gain and group distance ratio under all depths in Appendix.

\begin{table}[t!]
\setlength{\abovecaptionskip}{0.cm}
\setlength{\belowcaptionskip}{-0.1cm}
  \centering
  \caption{The highest accuracy (\%) and the accompanied optimal layers in the scenario with missing features. We calculate the average improvement achieved by DGN over each GNN framework.}
  \label{tab: test-compare-missing}
  \begin{tabular}{c|c|cc|cc|cc|cc|c}
    \toprule
    \multirow{2}*{\textbf{Model}} & \multirow{2}*{\textbf{Norm}} & \multicolumn{2}{c}{\textbf{Cora}} & \multicolumn{2}{c}{\textbf{Citeseer}} & \multicolumn{2}{c}{\textbf{Pubmed}} & \multicolumn{2}{c|}{\textbf{CoauthorCS}} & 
    \multirow{2}*{Improvement\%}\\
    \cline{3-10}
    & & Acc & \#K & Acc & \#K & Acc & \#K & Acc & \#K \\
    \hline
    \hline
    \multirow{4}*{GCN} & NN & $57.3$ & $3$ & $44.0$ & $6$ & $36.4$ & $4$ & $67.3$ & $3$ & $42.2$\\
    & BN & $71.8$ & $20$ & $45.1$ & $25$ & $70.4$ & $30$ & $82.7$ & $30$ & $5.2$\\
    & PN & $65.6$ & $20$ & $43.6$ & $25$ & $63.1$ & $30$ & $63.5$ & $4$ & $19.2$\\
    & DGN & $\bm{76.3}$ & $\bm{20}$ & $\bm{50.2}$ & $\bm{30}$ & $\bm{72.0}$ & $\bm{30}$ & $\bm{83.7}$ & $\bm{25}$ & - \\
    \hline
    \multirow{4}*{GAT} & NN & $50.1$ & $2$ & $40.8$ & $4$ & $38.5$ & $4$ & $63.7$ & $3$ & $51.0$\\
    & BN & $72.7$ & $5$ & $48.7$ & $5$ & $60.7$ & $4$ & $80.5$ & $6$ & $9.8$ \\
    & PN & $68.8$ & $8$ & $50.3$ & $6$ & $63.2$ & $20$ & $66.6$ & $3$ & $14.7$\\
    & DGN & $\bm{75.8}$ & $\bm{8}$ & $\bm{54.5}$ & $\bm{5}$ & $\bm{72.3}$ & $\bm{20}$ & $\bm{83.6}$ & $\bm{15}$ & - \\
    \hline
    \multirow{4}*{SGC} & NN & $63.4$ & $5$ & $51.2$ & $40$ & $63.7$ & $5$ & $71.0$ & $5$ & $20.1$\\
    & BN & $78.5$ & $20$ & $50.4$ & $20$ & $72.3$ & $50$ & $84.4$ & $20$ & $6.2$\\
    & PN & $73.4$ & $50$ & $58.0$ & $120$ & $75.2$ & $30$ & $80.1$ & $10$ & $4.5$ \\
    & DGN & $\bm{80.2}$ & $\bm{50}$ & $\bm{58.2}$ & $\bm{90}$ & $\bm{76.2}$ & $\bm{90}$ & $\bm{85.8}$ & $\bm{20}$ & - \\
\bottomrule
  \end{tabular}
\end{table}

\begin{figure}
\advance\leftskip 0cm
\setlength{\belowcaptionskip}{-0.45cm}
    \includegraphics[width=.27\textwidth]{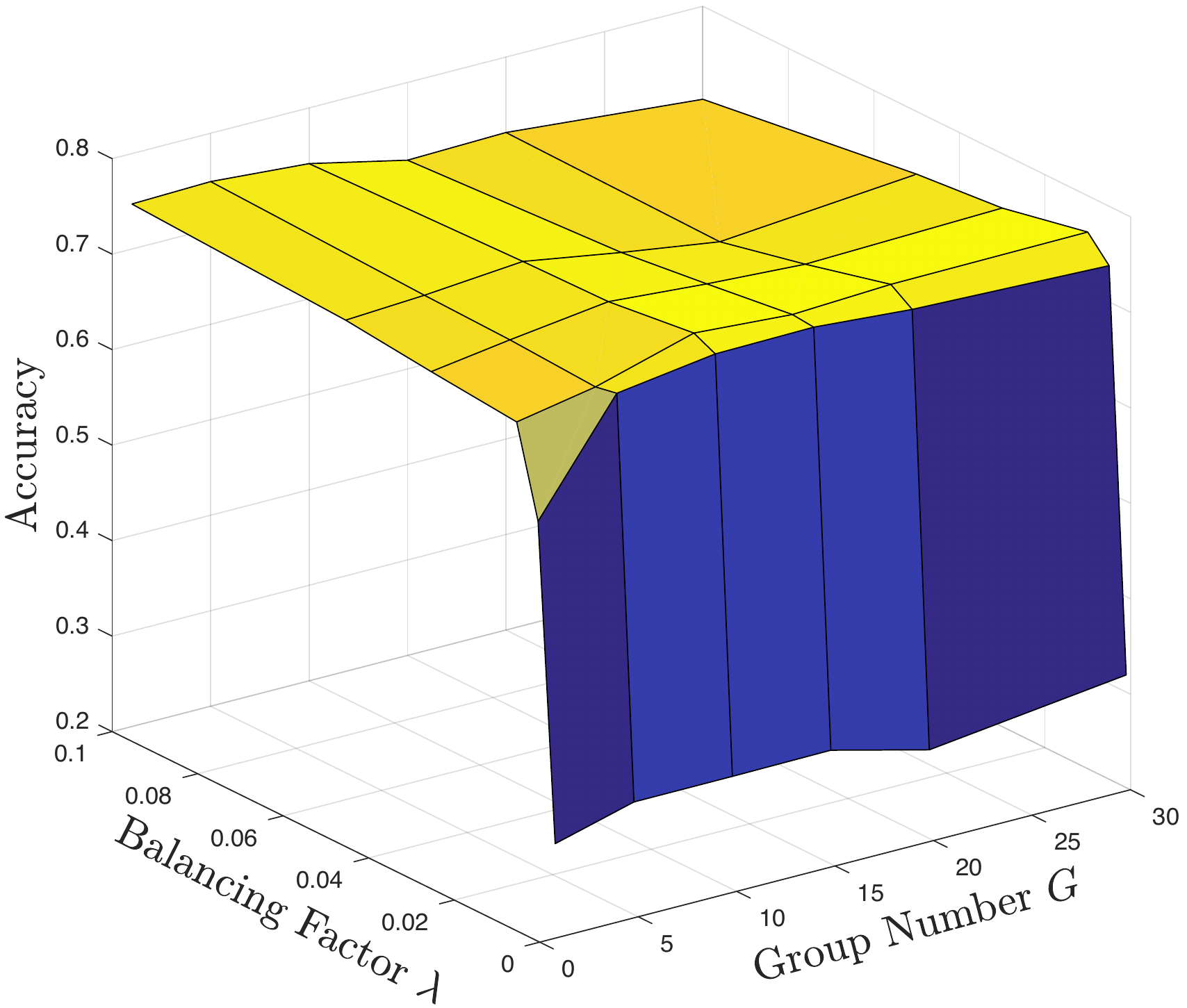} \hfill
    \includegraphics[width=.31\textwidth]{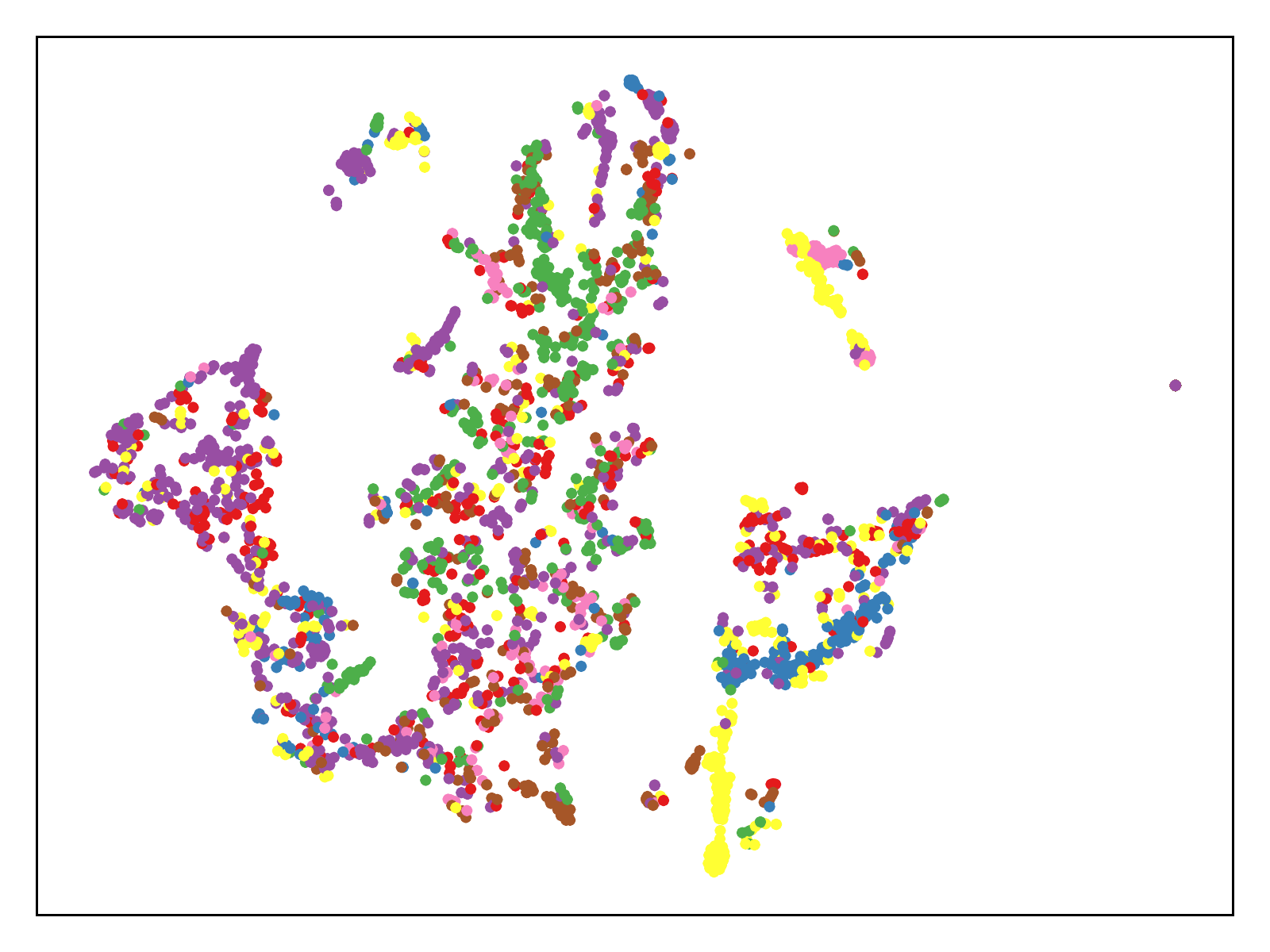}\hfill
    \includegraphics[width=.31\textwidth]{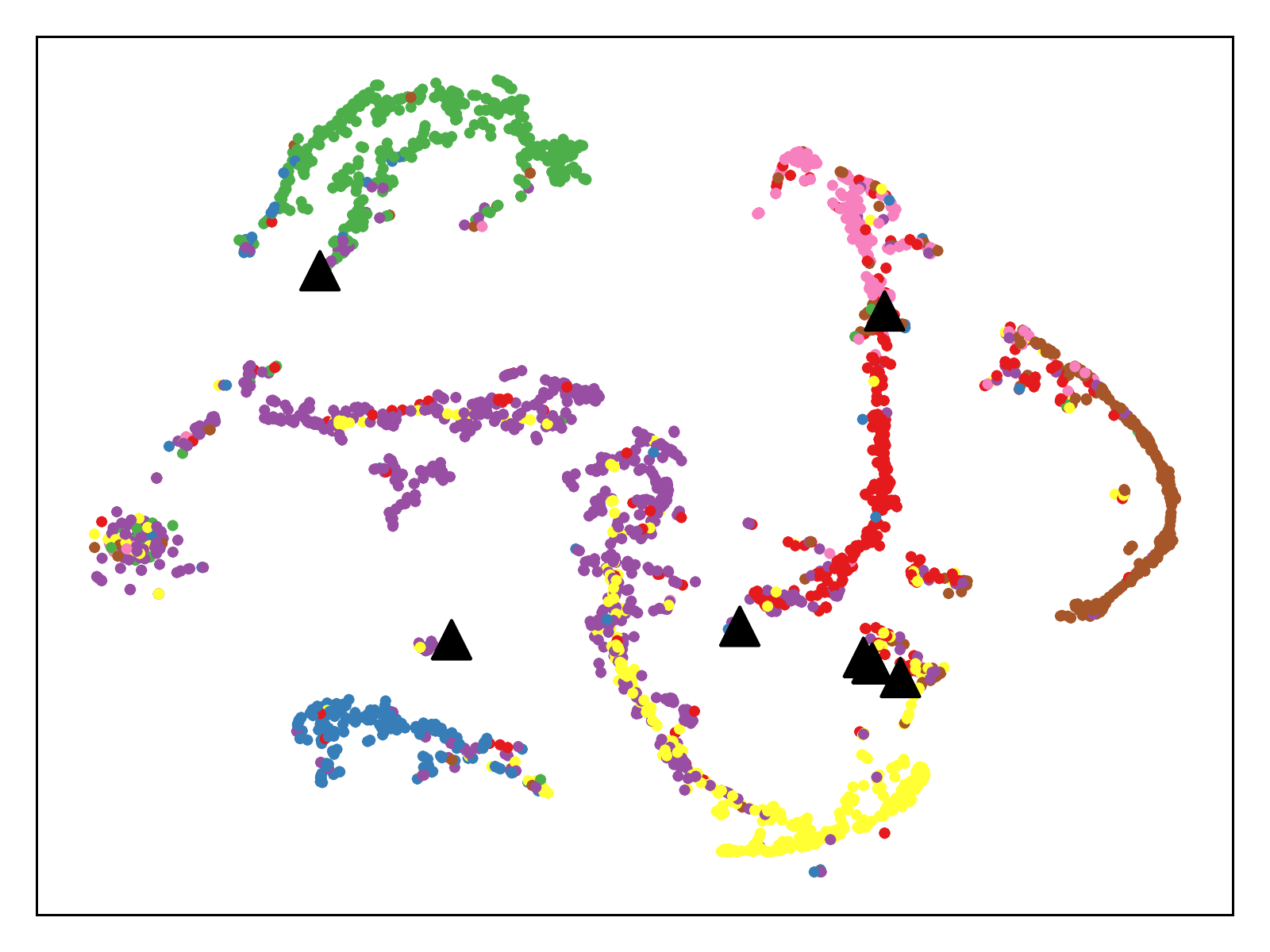}
    \caption{\textbf{Left}: Test accuracies of GCN with $20$ layers on Cora with missing features, where hyperparameters $G$ and $\lambda$ are studied. \textbf{Middle}: Node representation visualization for GCN without normalization and with $K=20$. \textbf{Right}: Node representation visualization for GCN with DGN layer and $K=20$ (node colors represent classes, and black triangles denote the running means of groups).}
    \label{fig: hyper-vis}
    
\end{figure}

% \paragraph{Hyperparameter studies.} 
\textbf{Hyperparameter studies.} 
We study the impact of hyperparameters, group number $G$ and balancing factor $\lambda$, on DGN in order to answer research question \textbf{Q3}. Over the GCN framework associated with $20$ convolutional layers, we evaluate DGN by considering $G$ and $\lambda$ from sets $[1, 5, 10, 15, 20, 30]$ and $[0.001, 0.005, 0.01, 0.03, 0.05, 0.1]$, respectively. The left part in Figure \ref{fig: hyper-vis} presents the test accuracy for each hyperparameter combination. We observe that: (i) The model performance is damaged greatly when $\lambda$ is close to zero (e.g., $\lambda = 0.001$). In this case, group normalization contributes slightly in DGN, resulting in over-smoothing in the GCN model. (ii) Model performance is not sensitive to the value of $G$, and an appropriate $\lambda$ value could be tuned to optimize the trade-off between instance gain and group normalization. It is because DGN learns to use the appropriate number of groups by end-to-end training. In particular, some groups might not be used as shown in the right part of Figure \ref{fig: hyper-vis}, at which only $6$ out of $10$ groups (denoted by black triangles) are adopted.  (iii) Even when $G = 1$, DGN still outperforms BN by utilizing the self-preserved component to achieve an accuracy of $74.7\%$, where $\lambda = 0.1$. Via increasing the group number, the model performance could be further improved, e.g., the accuracy of $76.3\%$ where $G = 10$ and $\lambda = 0.01$.

% \paragraph{Node representation visualization.} 
\textbf{Node representation visualization.}
We investigate how DGN clusters nodes into different groups to tackle the over-smoothing issue. The middle and right parts of Figure \ref{fig: hyper-vis} visualize the node representations achieved by GCN models without normalization tool and with the DGN approach, respectively. It is observed that the node representations of different classes mix together when the layer number reaches $20$ in the GCN model without normalization. In contrast, our DGN method softly assigns nodes into a series of groups, whose running means at the corresponding normalization modules are highlighted with black triangles. Through normalizing each group independently, the running means are separated to improve inter-group distances and disentangle node representations. In particular, we notice that the running means locate at the borders among different classes (e.g., the upper-right triangle at the border between red and pink classes). That is because the soft assignment may cluster nodes of two or three classes into the same group. Compared with batch or pair normalization, the independent normalization for each group only includes a few classes in DGN. In this way, we relieve the representation noise from other node classes during normalization, and improve the group distance ratio as illustrated in Appendix.

\section{Conclusion}
%\textcolor{red}{study} how to measure the over-smoothing issue of GNNs from the perspective of nodes' groups, and 
In this paper, we propose two over-smoothing metrics based on graph structures, i.e., group distance ratio and instance information gain. By inspecting GNN models through the lens of these two metrics, we present a novel normalization layer, DGN, to boost model performance against over-smoothing. It normalizes each group of similar nodes independently to separate node representations of different classes. Experiments on real-world classification tasks show that DGN greatly slowed down performance degradation by alleviating the over-smoothing issue. DGN enables us to explore deeper GNNs and achieve higher performance in analyzing attributed networks and the scenario with missing features. Our research will facilitate deep learning models for potential graph applications.

% \newpage

\section*{Broader Impact}
The successful outcome of this work will lead to advances in building up deep graph neural networks and dealing with complex graph-structured data. The developed metrics and algorithms have an immediate and strong impact on a number of fields, including (1) \textit{Over-smoothing Quantitative Analysis}: GNN models tend to result in the over-smoothing issue with the increase in the number of layers. During the practical development of deeper GNN models, the proposed instance information gain and group distance ratio effectively indicate the over-smoothing issue, in order to push the model exploration toward a good direction. (2) \textit{Deep GNN Modeling}: The proposed differentiable group normalization tool successfully tackles the over-smoothing issue and enables the modeling of deeper GNN variants. It encourages us to fully unleash the power of deep learning in processing the networked data. (3) \textit{Real-world Network Analytics Applications}: The proposed research will broadly shed light on utilizing deep GNN models in various applications, such as social network analysis, brain network analysis, and e-commerce network analysis. For such complex graph-structured data, deep GNN models can exploit the multi-hop neighborhood information to boost the task performance.

\bibliographystyle{unsrt}
\bibliography{reference}

\newpage

\appendix 
\section{Dataset Statistics}
For fair comparison with previous work, we perform the node classification task on four benchmark datasets, including Cora, Citeseer, Pubmed~\cite{yang2016revisiting}, and CoauthorCS~\cite{shchur2018pitfalls}. They have been widely adopted to study the over-smoothing issue in GNNs~\cite{hou2020measuring, chen2019measuring, zhao2019pairnorm, li2018deeper, rong2020dropedge}. The detailed statistics are listed in Table~\ref{tab:dataset}. To further illustrate that the normalization techniques could enable deeper GNNs to achieve better performance, we 
apply them to a more complex scenario with missing features. For these four benchmark datasets, we create the corresponding scenarios by removing node features in both validation and testing sets.

\begin{table}[h]
\centering
\setlength{\tabcolsep}{15pt}
  \caption{Dataset statistics on Cora, Citeseer, Pubmed, and CoauthorCS.}
  \label{tab:dataset}
  \begin{tabular}{c|cccc}
    \toprule
     & Cora & Citeseer & Pubmed & CoauthorCS\\
    \midrule
    \#Nodes & $2708$ & $3327$ & $19717$ & $18333$ \\
    \#Edges & $5429$ & $4732$ & $44338$ & $81894$ \\
    \#Features & $1433$ & $3703$ & $500$ & $6805$ \\
    \#Classes& $7$ & $6$ & $3$ & $15$ \\
    \#Training Nodes & $140$ & $120$ & $60$ & $600$ \\
    \#Validation Nodes & $500$ & $500$ & $500$ &  $2250$ \\
    \#Testing Nodes & $1000$ & $1000$ & $1000$ & $15483$  \\
  \bottomrule
\end{tabular}
\end{table}

\section{Running Environment}
All the GNN models and normalization approaches are implemented in PyTorch, and tested on a machine with 24 Intel(R) Xeon(R) CPU E5-2650 v4 @ 2.20GB processors, GeForce GTX-1080 Ti 12 GB GPU, and 128GB memory size. We implement the group normalization in a parallel way. Thus the practical time cost of our DGN is comparable to that of traditional batch normalization. 

\section{GNN Models}
We test over three general GNN models to illustrate the over-smoothing issue, including graph convolutional networks (GCN)~\cite{kipf2016semi}, graph attention networks (GAT)~\cite{velickovic2017graph} and simple graph convolution (SGC) networks~\cite{wu2019simplifying}. We list their neighbor aggregation functions in Table~\ref{tab:GNN_models}.

\begin{table}[h]
\centering
\setlength{\tabcolsep}{20pt}
  \caption{Neighbor aggregation function at a graph convolutional layer for GCN, GAT and SGC.}
  \label{tab:GNN_models}
  \begin{tabular}{c|c}
    \toprule
     Model & Neighbor aggregation function \\
    \midrule
    \hline
    GCN & $h^{(k)}_v = \mathrm{ReLU}(\sum_{v'\in \mathcal{N}(v) \cup \{v\}} \frac{1}{\sqrt{(|\mathcal{N}(v)|+1)\cdot(|\mathcal{N}(v')|+1)}} W^{(k)} h_{v'}^{(k-1)})$ \\
    \hline
    GAT & $h^{(k)}_v = \mathrm{ReLU}(\sum_{v'\in \mathcal{N}(v) \cup \{v\}} a_{vv'}^{(k)} W^{(k)} h_{v'}^{(k-1)})$ \\
    \hline
    SGC & $h^{(k)}_v = \sum_{v'\in \mathcal{N}(v) \cup \{v\}} \frac{1}{\sqrt{(|\mathcal{N}(v)|+1)\cdot(|\mathcal{N}(v')|+1)}} h_{v'}^{(k-1)}$ \\
  \bottomrule
\end{tabular}
\end{table}

Considering the message passing strategy as shown by Eq. (1) in the main manuscript, we explain the key properties of GCN, GAT and SGC as follows. GCN merges the information from node itself and its neighbors weighted by vertices' degrees, where $a^{(k)}_{vv'} = 1. / \sqrt{(|\mathcal{N}(v)|+1) \cdot (|\mathcal{N}(v')|+1)}$. Functions $\mathrm{AGG}$ and $\mathrm{COM}$ are realized by a summation pooling. The activation function of ReLU is then applied to non-linearly transform the latent embedding. Based on GCN, GAT uses an additional attention layer to learn link weight $a^{(k)}_{vv'}$. GAT aggregates neighbors with the trainable link weights, and achieves significant improvements in a variety of applications. SGC is simplified from GCN by removing all trainable parameters $W^{(k)}$ and nonlinear activations between successive layers. It has been empirically shown that these simplifications do not negatively impact classification accuracy, and even relive the problems of over-fitting and vanishing gradients in deeper models. 

\section{Normalization Baselines}
Batch normalization is first applied between the successive convolutional layers in CNNs~\cite{ioffe2015batch}. It is extended to graph neural networks to improve node representation learning and generalization~\cite{dwivedi2020benchmarking}. Taking embedding matrix $H^{(k)}$ as input after each layer, batch normalization scales the node representations using running mean and variance, and generates a new embedding matrix for the next graph convolutional layer. Formally, we have:
$$
    \tilde H^{(k)} = \gamma (\frac{H^{(k)} - \mu}{\sigma}) + \beta \in \mathbb{R}^{n\times d^{(k)}}.
$$
$\mu$ and $\sigma$ denote the vectors of running mean and standard deviation, respectively; $\gamma$ and $\beta$ denote the trainable scale and shift vectors, respectively. Recently, pair normalization has been proposed to tackle the over-smoothing issue in GNNs, targeting at maintaining the average node pair distance over a graph~\cite{zhao2019pairnorm}. Pair normalization is a simplifying realization of batch normalization by removing the trainable $\gamma$ and $\beta$. In this work, we augment each graph convolutional layer via appending a normalization module, in order to validate the effectiveness of normalization technique in relieving over-smoothing and enabling deeper GNNs.

\section{Hyperparameter Tuning in DGN}
The balancing factor, $\lambda$, is crucial to determine the trade-off between input feature preservation and group normalization in DGN. It needs to be tuned carefully as GNN models increase the number of layers. To be specific, we consider the candidate set $\{5\cdot 10^{-4}, 1\cdot 10^{-3}, 2\cdot 10^{-3}, 3\cdot 10^{-3}, 5\cdot 10^{-3}, 1\cdot 10^{-2}, 2\cdot 10^{-2}, 3\cdot 10^{-2}, 5\cdot 10^{-2}\}$. For each specific model, we use a few epochs to choose the optimal $\lambda$ on the validation set, and then evaluate it on the testing set. We observe that the value of $\lambda$ tends to be larger in the model accompanied with more graph convolutional layers. That is because the over-smoothing issue gets worse with the increase in layer number. The group normalization is much more required to separate the node representations of different classes.

% selected from $\{1\cdot 10^{-3}, 5\cdot 10^{-3}, 1\cdot 10^{-2}, 3\cdot 10^{-2}\}$, based on the criterion of using larger value in more deeper model to attend on representation normalization and relieve over-smoothing issue. The detail configurations of $G$ and $\lambda$ are listed in Appendix for each dataset.

\section{Instance Information Gain}
In this work, we adopt kernel-density estimators (KDE), one of the common non-parametric approaches, to estimate the mutual information between input feature and representation vector~\cite{kolchinsky2017estimating, kolchinsky2019nonlinear}. A key assumption in KDE is that the input feature (or output representation vector) of neural networks is distributed as a mixture of Gaussians. Since a neural network is a deterministic function of the input feature after training, the mutual information would be infinite without such assumption. In the following, we first formally define the Gaussian assumption, input probability distribution and representation probability distribution, and then present how to obtain the instance information gain 
based on the mutual information metric.  

\paragraph{Gaussian assumption.} In the graph signal processing, it is common to assume that the collected input feature contains both true signal and noise. In other word, we have the input feature as follows: $x_v = \bar{x}_v + \epsilon_x$. $\bar{x}_v$ denotes the true value, and $\epsilon_x \sim \mathcal{N}(0, \sigma^2\bm{I})$ denotes the added Gaussian noise with variance $\sigma^2$. Therefore, input feature $x_v$ is a Gaussian variable centered on its true value.

\paragraph{Input probability distribution.} We treat the empirical distribution of input samples as true distribution. Given a dataset accompanied with $n$ samples, we have a series of input features $\{x_1, \cdots, x_n\}$ for all the samples. Each node feature is sampled with probability $1 / |\mathcal{V}|$ following the empirical uniform distribution. Let $|\mathcal{V}|$ denotes the number of samples, and let $\mathcal{X}$ denote the random variable of input features. Based on the above Gaussian assumption, probability  $P_{\mathcal{X}}(x_v)$ of input feature $x_v$ is obtained by the product of $1 / |\mathcal{V}|$ with Gaussian probability centered on true value $\bar x_v$.

\paragraph{Representation probability distribution.} Let $\mathcal{H}$ denote the random variable of node representations. To obtain probability $P_{\mathcal{H}}(h_v)$ of continuous vector $h_v$, a general approach is to bin and transform $\mathcal{H}$ into a new discrete variable. However, with the increasing dimensions of $h_v$, it is non-trivial to statistically count the frequencies of all possible discrete values. Considering the task of node classification, the index of largest element along vector $h_v\in \mathbb{R}^{C\times 1}$ is regarded as the label of a node. We propose a new binning approach that labels the whole vector $h_v$ with the largest index $z_v$. In this way, we only have $C$ classes of discrete values to facilitate the frequency counting. To be specific, let $\bf P_c$ denote the number of representation vectors whose indexes $z_v = c$. The probability of a discrete variable with class $c$ is given by: $p_c = P_{\mathcal{H}}(z_v = c) = \frac{\bf P_c}{\sum_{l=1}^{C}\bf P_l}$.

\paragraph{Mutual information calculation.} Based on KDE approach, a lower bound of mutual information between input feature and representation vector can be calculated as:
$$
\begin{array}{rl}
    G_{\mathrm{Ins}} = I(\mathcal{X}; \mathcal{H}) & = \sum_{x_v\in \mathcal{X}, h_v \in \mathcal{H}} P_{\mathcal{XH}}(x_v, h_v) \log \frac{P_{\mathcal{XH}}(x_v, h_v)}{P_{\mathcal{X}}(x_v)P_{\mathcal{H}}(h_v)} \\
    \\
    & = H({\mathcal{X}}) - H({\mathcal{X}} | {\mathcal{H}}) \\
    \\
    & \geq -\frac{1}{|\mathcal{V}|}\sum_{i} \log{\frac{1}{|\mathcal{V}|}} \sum_j\exp(-\frac{1}{2} \frac{||x_i - x_j||_2^2}{4\sigma^2}) \\
    & \quad -\sum_{c=1}^C p_c [-\frac{1}{\bf P_c}\sum_{i, z_i=c} \log{\frac{1}{\bf P_c}} \sum_{j, z_j=c}\exp(-\frac{1}{2} \frac{||x_i - x_j||_2^2}{4\sigma^2})].
\end{array}
$$
The sum over $i, z_i = c$ represents a summation over
all the input features whose representation vectors are labeled with $z_i = c$. $P_{\mathcal{XH}}(x_v, h_v)$ denotes the joint probability of $x_v$ and $h_v$. The effectiveness of $G_{\mathrm{Ins}}$ in measuring mutual information between input feature and node representation has been demonstrated in the experimental results. As illustrated in Figures \ref{fig:all_Cora_1}-\ref{fig:all_CoauthorCS}, $G_{\mathrm{Ins}}$ decreases with the increasing number of graph convolutional layers. This practical observation is in line with the human expert knowledge about neighbor aggregation strategy in GNNs. The neighbor aggregation function as shown in Table~\ref{tab:GNN_models} is in fact a low-passing smoothing operation, which mixes the input feature of a node with those of its neighbors gradually. At the extreme cases where $K=30$ or $120$, we find that $G_{\mathrm{Ins}}$ approaches to zero in GNN models without normalization. The loss of informative input feature leads to the dropping of node classification accuracy. However, our DGN keeps the input information during graph convolutions and normalization to some extent, resulting in the largest $G_{\mathrm{Ins}}$ compared with the other normalization approaches.

\section{Performance Comparison on Attributed Graphs}
\begin{figure}[htbp]
    \begin{minipage}[t]{\linewidth}
		\centering
		\includegraphics[width=\linewidth]{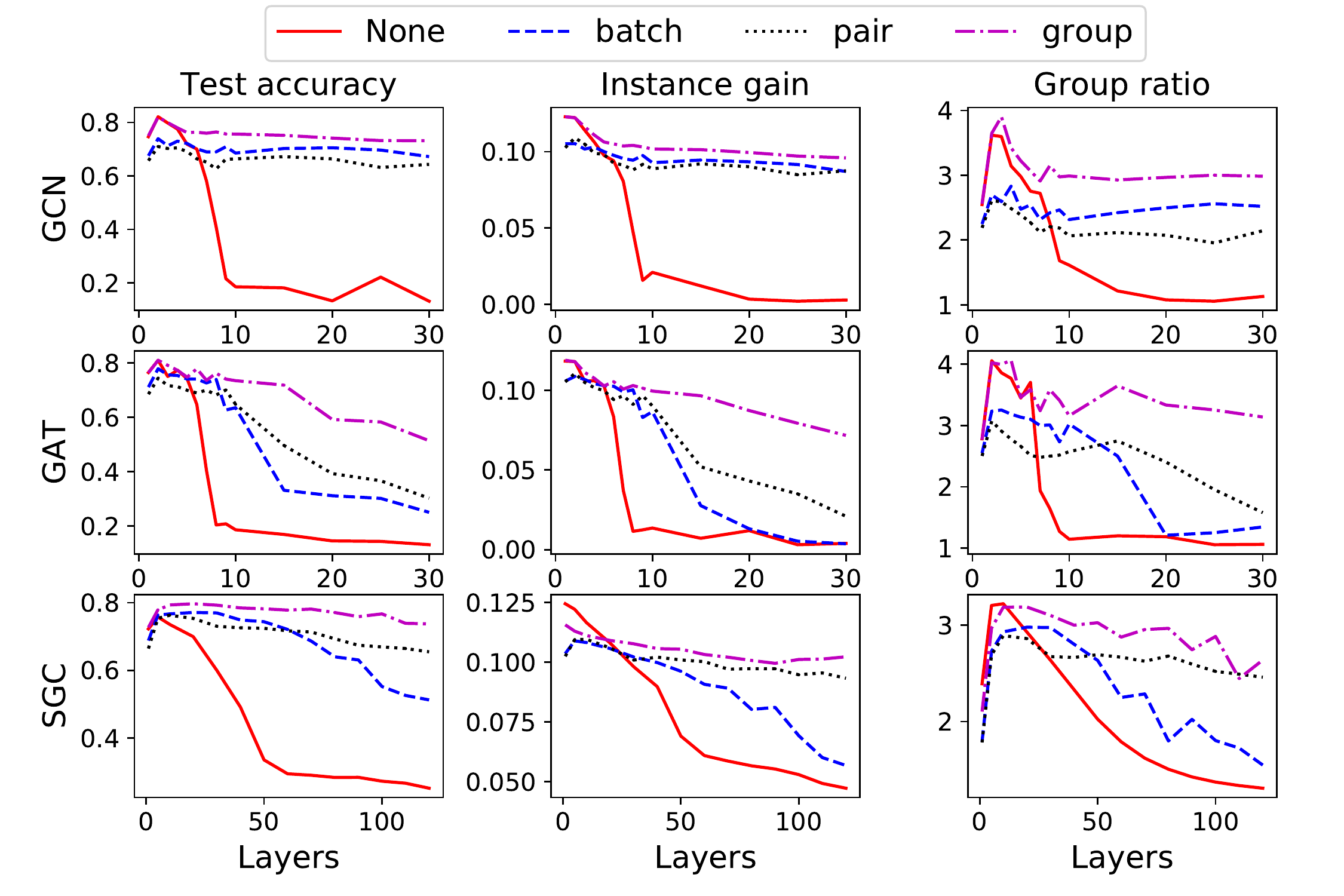}
		\caption{The test accuracy, instance information gain, and group distance ratio in attributed Cora. We compare differentiable group normalization with none, batch and pair normalizations.}
		\label{fig:all_Cora_1}
	\end{minipage}
	
	\vspace{0.5cm}
	\begin{minipage}[t]{\linewidth}
		\centering
		\includegraphics[width=\linewidth]{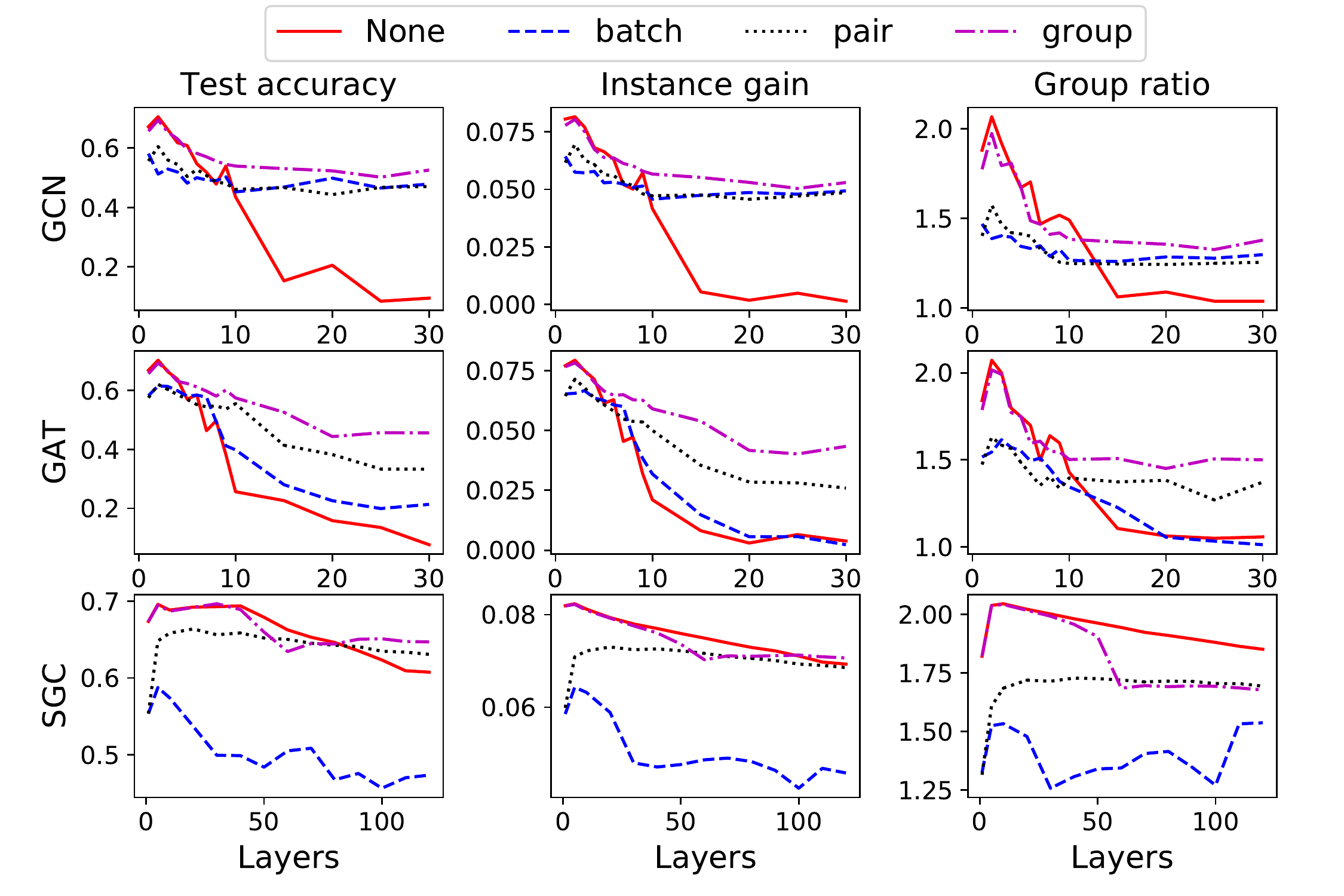}
		\caption{The test accuracy, instance information gain, and group distance ratio in attributed Citeseer. We compare differentiable group normalization with none, batch and pair normalizations.}
		\label{fig:all_Citeseer}
	\end{minipage}
\end{figure}

\begin{figure}[htbp]
	\begin{minipage}[t]{\linewidth}
		\centering
		\includegraphics[width=\linewidth]{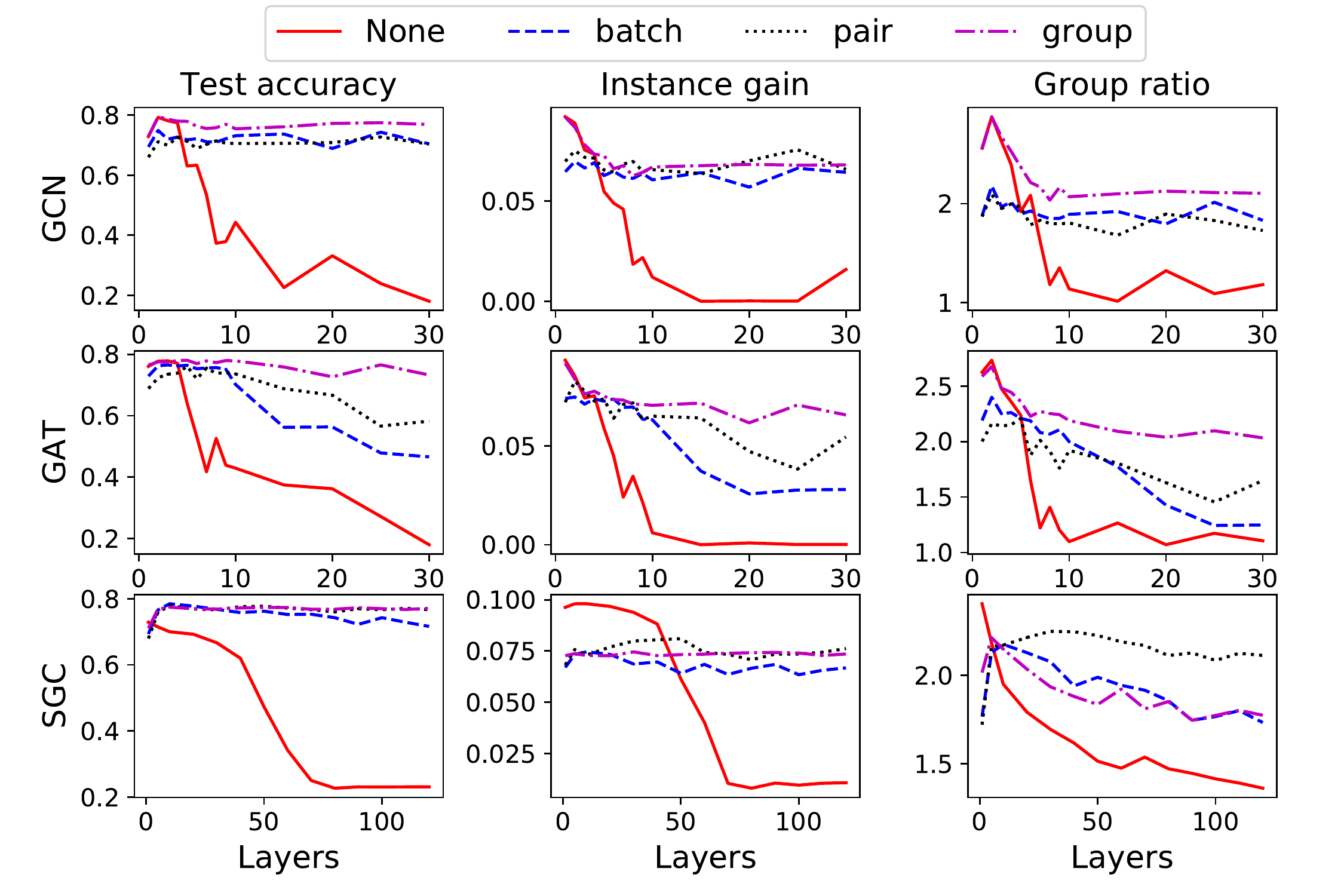}
		\caption{The test accuracy, instance information gain, and group distance ratio in attributed Pubmed. We compare differentiable group normalization with none, batch and pair normalizations.}
		\label{fig:all_Pubmed}
	\end{minipage}
	
	\vspace{0.5cm}
	\begin{minipage}[t]{\linewidth}
	    \centering
		\includegraphics[width=\linewidth]{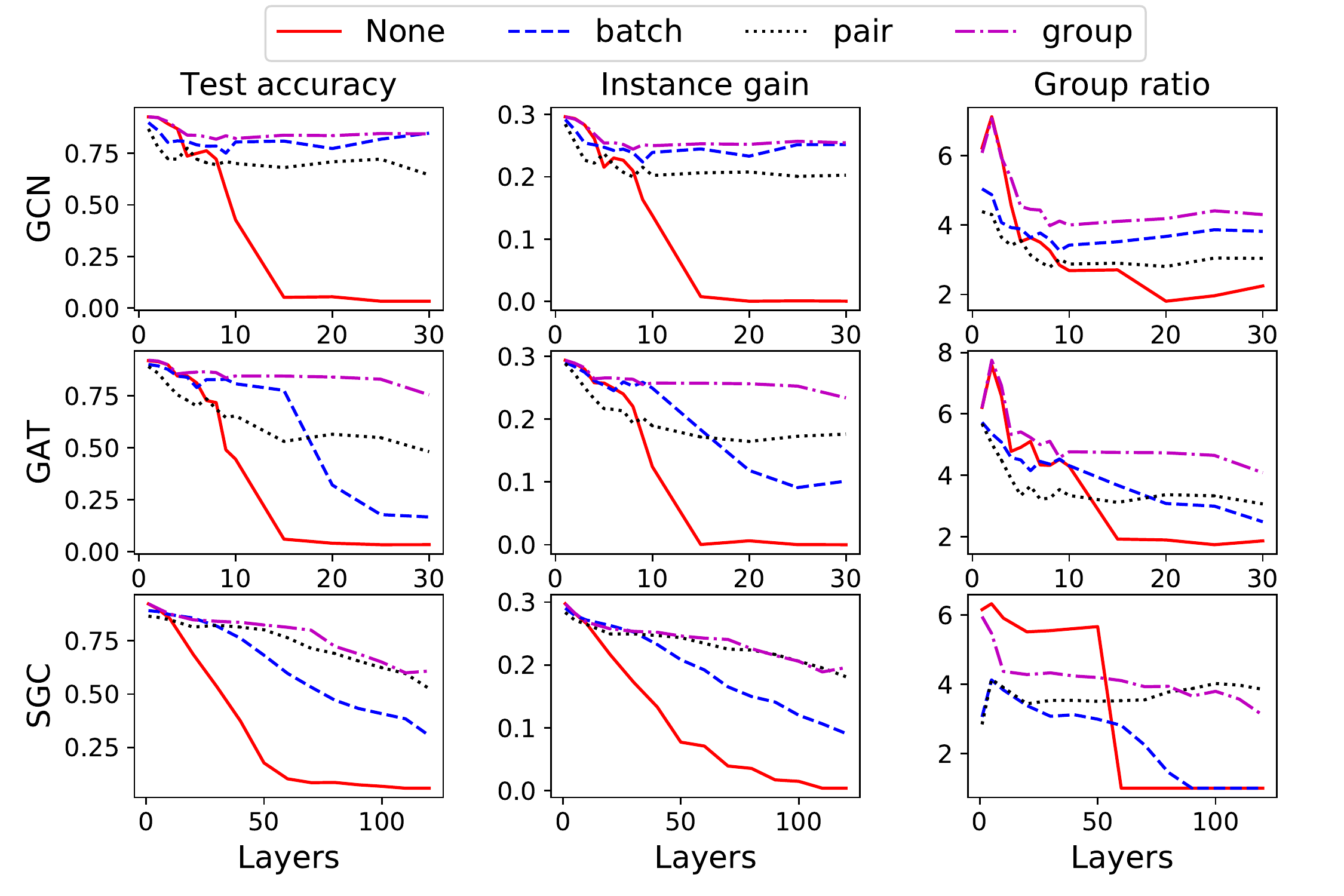}
		\caption{The test accuracy, instance information gain, and group distance ratio in attributed CoauthorCS. We compare differentiable group normalization with none, batch and pair normalizations.}
		\label{fig:all_CoauthorCS}
	\end{minipage}
\end{figure}

In this section, we report the model performances in terms of test accuracy, instance information gain and group distance ratio achieved on all the concerned datasets in Figures \ref{fig:all_Cora_1}-\ref{fig:all_CoauthorCS}. We make the following observations:
\begin{itemize}[wide=0pt, leftmargin=\dimexpr\labelwidth + 2\labelsep\relax]
    \item Comparing with other normalization techniques, our DGN generally slows down the dropping of test accuracy with the increase in layer number. Even for GNN models associated with a small number of layers (i.e., $G \leq 5$), DGN achieves the competitive performance compared with none normalization. The adoption of DGN module does not damage the model performance, and prevents model from suffering over-smoothing issue when GNN goes deeper.
    
    \item DGN achieves the larger or comparable instance information gains in all cases, especially for GAT models. That is because DGN keeps embedding matrix $H^{(k)}$ and prevents over-normalization within each group. The preservation of $H^{(k)}$ saves input features to some extent after each layer of graph convolutions and normalization. In an attributed graph, the improved preservation of informative input features in the final representations will significantly facilitate the downstream node classification. Furthermore, such preservation is especially crucial for GNN models with a few layers, since the over-smoothing issue has not appeared. 
    
    \item DGN normalizes each group of node representations independently to generally improve the group distance ratio, especially for models GCN and GAT. A larger value of group distance ratio means that the node representation distributions from all groups are disentangled to address the over-smoothing issue. Although the ratios of DGN are smaller than those of pair normalization in some cases upon SGC framework, we still achieve the largest test accuracy. That may be because the intra-group distance in DGN is much smaller than that of pair normalization. A small value of intra-group distance would facilitate the node classification within the same group. We will further compare the intra-group distance in scenarios with missing features in the following experiments.
\end{itemize}

\newpage

\section{Performance Comparison in Scenarios with Missing Features}
\begin{figure}[htbp]
	\begin{minipage}[t]{\linewidth}
		\centering
		\includegraphics[width=\linewidth]{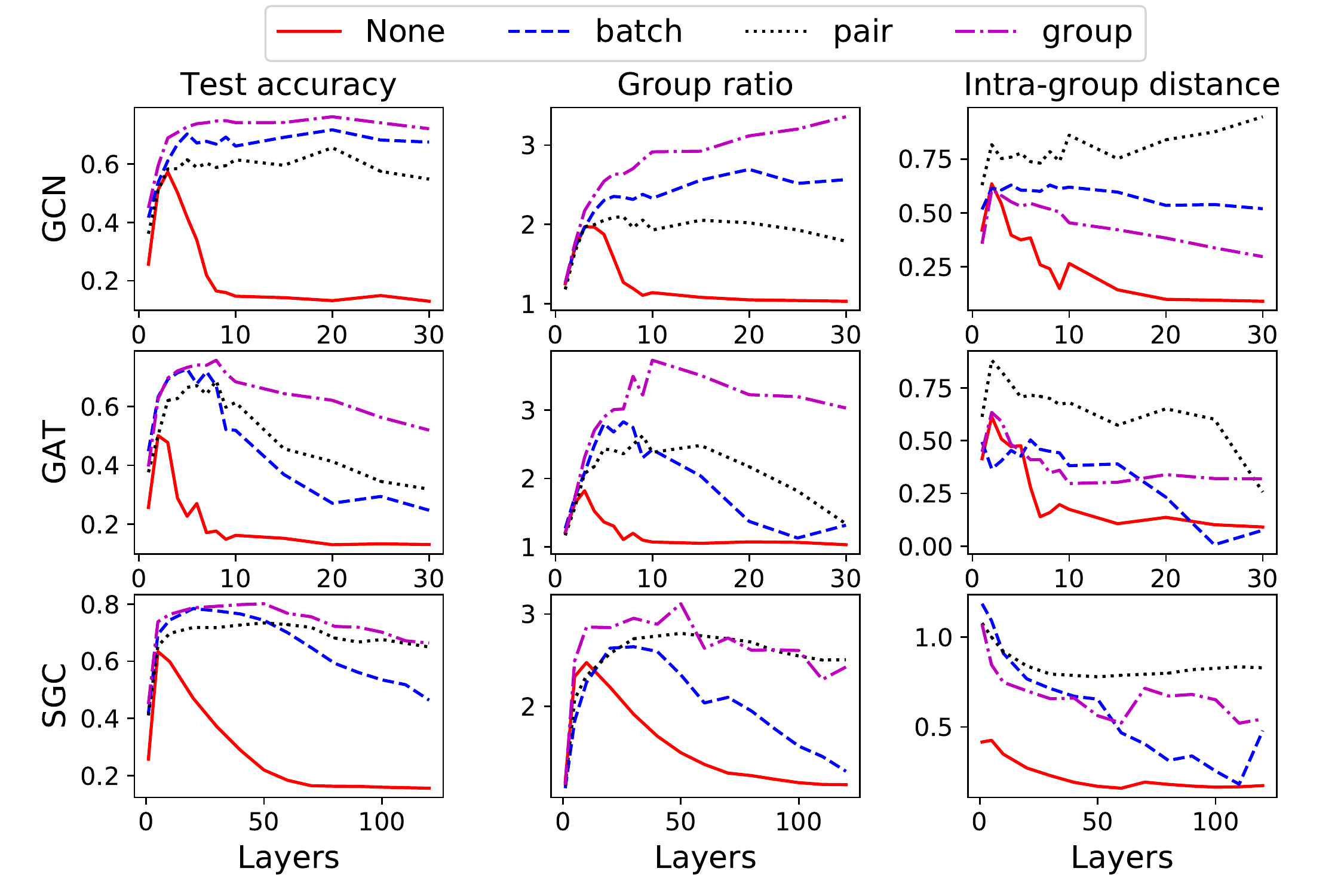}
		\caption{The test accuracy, group distance ratio and intra-group distance in Cora with missing features. We compare differentiable group normalization with none, batch and pair normalizations.}
		\label{fig:all_plain_Cora}
	\end{minipage}
	
	\vspace{0.5cm}
	\begin{minipage}[t]{\linewidth}
		\centering
		\includegraphics[width=\linewidth]{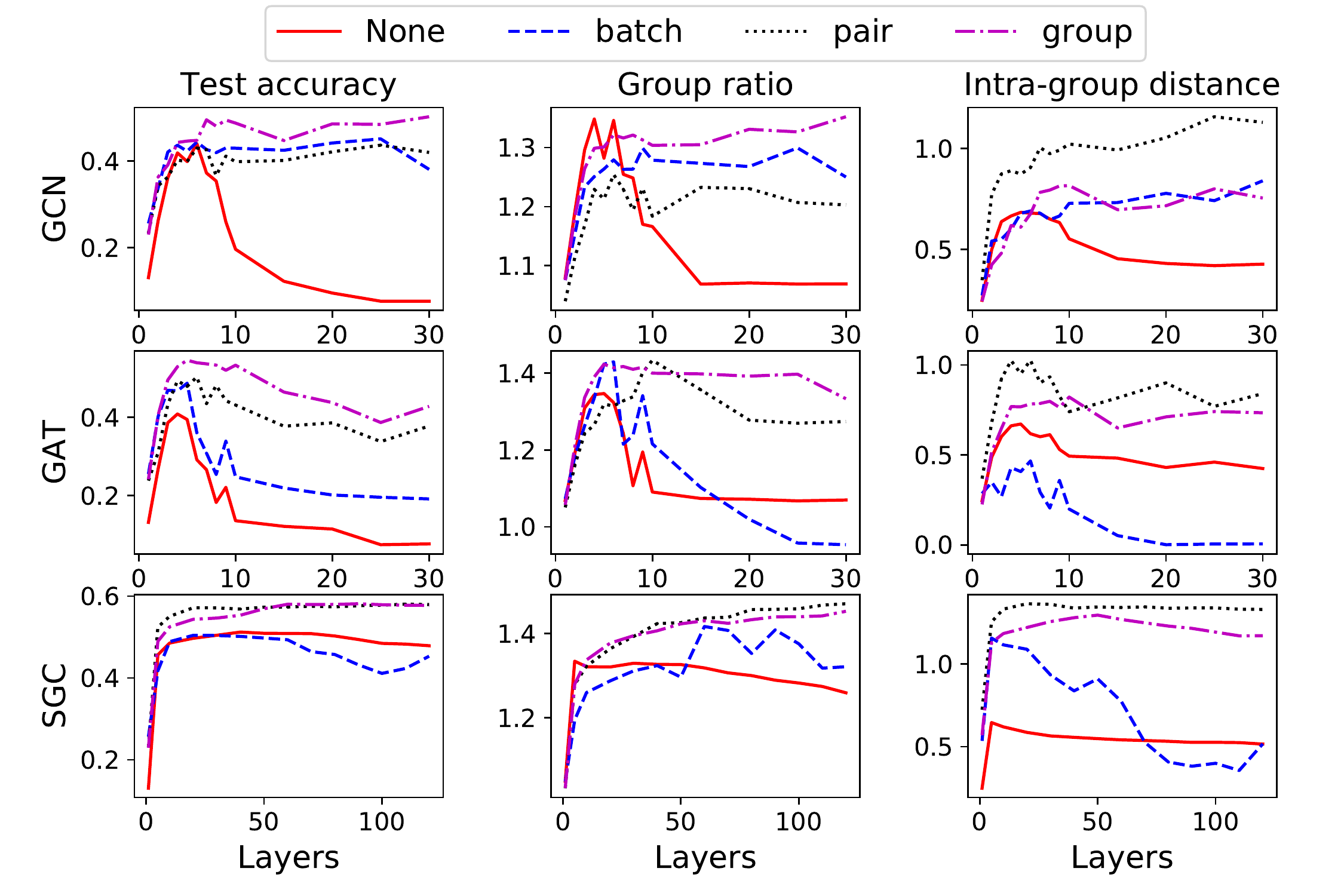}
		\caption{The test accuracy, group distance ratio and intra-group distance in Citeseer with missing features. We compare differentiable group normalization with none, batch and pair normalizations.}
		\label{fig:all_plain_Citeseer}
	\end{minipage}
\end{figure}

\begin{figure}[htbp]
	\begin{minipage}[t]{\linewidth}
		\centering
		\includegraphics[width=\linewidth]{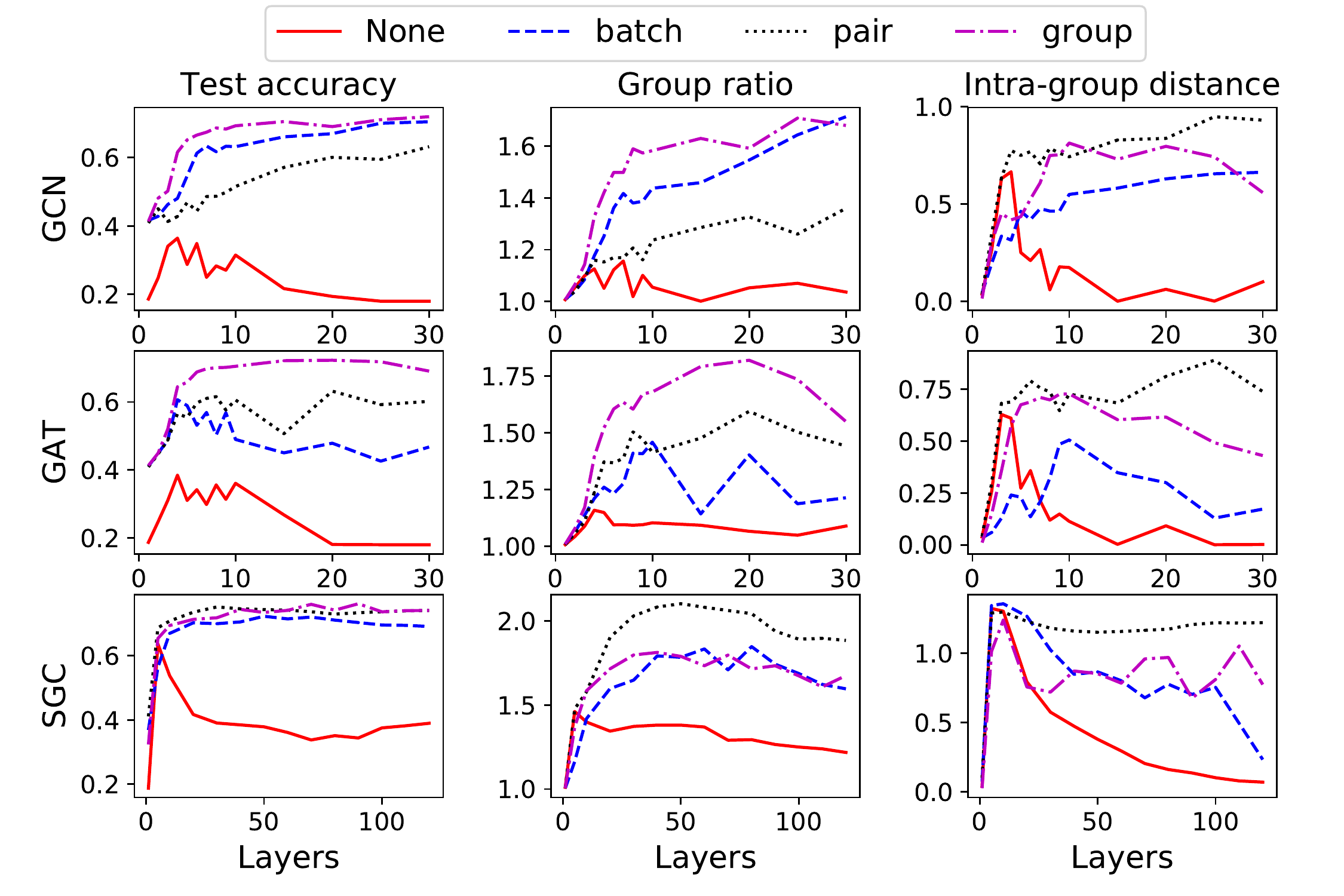}
		\caption{The test accuracy, group distance ratio and intra-group distance in Pubmed with missing features. We compare differentiable group normalization with none, batch and pair normalizations.}
		\label{fig:all_plain_Pubmed}
	\end{minipage}
	
	\vspace{0.5cm}
	\begin{minipage}[t]{\linewidth}
		\centering
		\includegraphics[width=\linewidth]{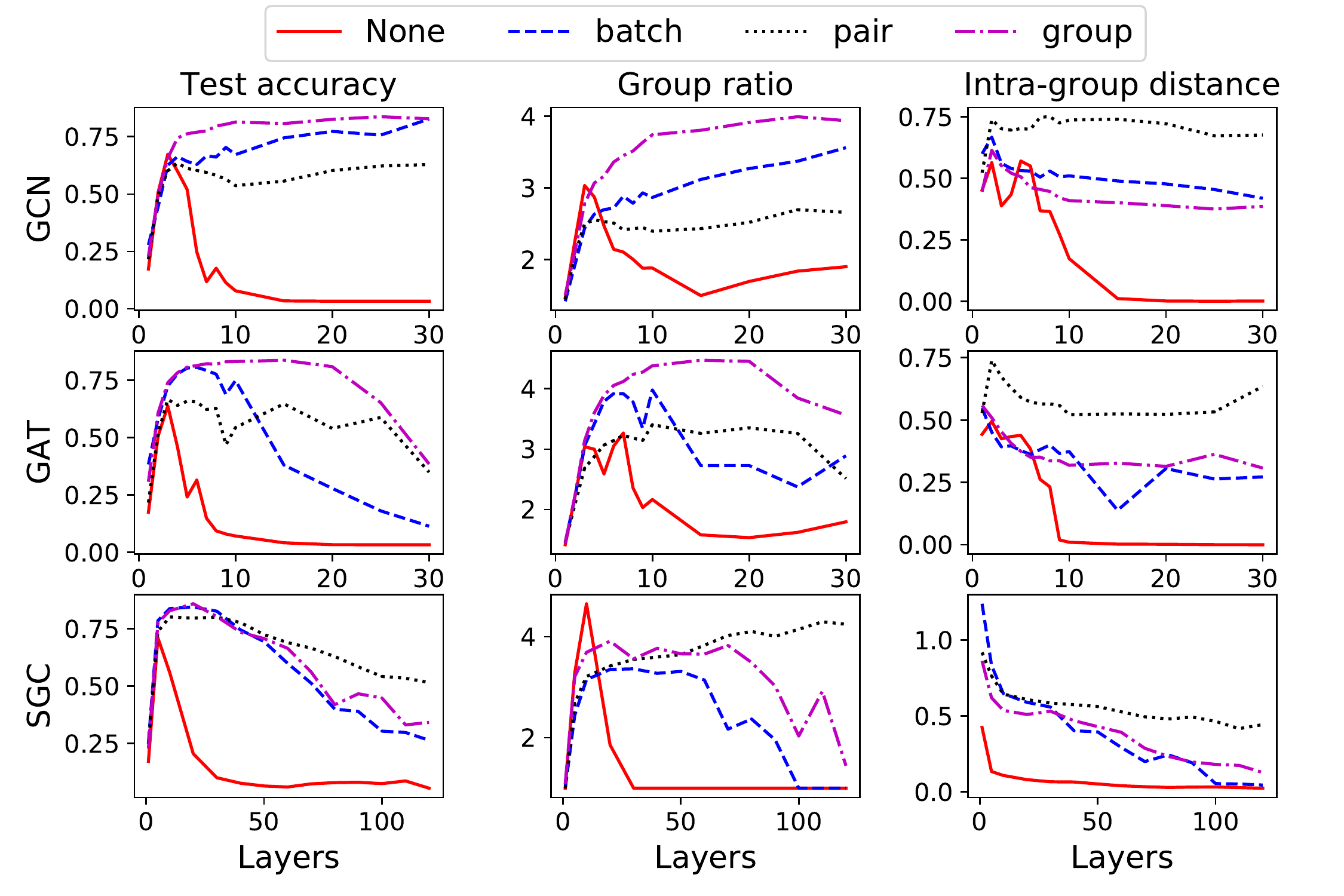}
		\caption{The test accuracy, group distance ratio and intra-group distance in CoauthorCS with missing features. We compare differentiable group normalization with none, batch and pair normalizations.}
		\label{fig:all_plain_CoauthorCS}
	\end{minipage}
\end{figure}

In this section, we report the model performances in terms of test accuracy, group distance ratio and intra-group distance achieved in scenarios with missing features in Figures \ref{fig:all_plain_Cora}-\ref{fig:all_plain_CoauthorCS}. The intra-group distance is calculated by node pair distance averaged within the same group. Its mathematical expression is given by the denominator of Equation (3) in the main manuscript. We make the following observations:
\begin{itemize}[wide=0pt, leftmargin=\dimexpr\labelwidth + 2\labelsep\relax]
    \item DGN achieves the largest test accuracy by exploring the deeper neural architecture with a larger number of graph convolutional layers. In the scenarios with missing features, GNN model relies highly on the neighborhood structure to classify nodes. DGN enables the deeper GNN model to exploit neighborhood structure with multiple hops away, and at the same time relieves the over-smoothing issue. 
    \item Comparing with other normalization techniques, DGN generally improves the group distance ratio to relieve over-smoothing issue. Although in some cases the ratios are smaller than those of pair normalization upon SGC framework, we still achieve the comparable or even better test accuracy. That is because DGN has a smaller intra-group distance to facilitate node classification within the same group, which is analyzed in the followings. 
    \item DGN obtains an appropriate intra-group distance to optimize the node classification task. While the over-smoothing issue results in an extremely-small distance in the model without normalization, a larger one in pair normalization leads to the inaccurate node classification within each group. That is because the pair normalization is designed to maintain the distance between each pair of nodes, no matter whether they locate in the same class group or not. The divergence of node representations in a group prevents a downstream classifier to assign them the same class label. 
\end{itemize}

\end{document}